\documentclass[runningheads]{llncs}
\usepackage{graphicx}
\usepackage{tikz}
\usepackage{comment} 
\usepackage{amsmath,amssymb} % define this before the line numbering.
\usepackage{color}

\usepackage{hyperref}

\usepackage{tabularx}

\usepackage{multirow}
\usepackage{booktabs}
\usepackage{colortbl}  % color tab
\definecolor{Gray}{gray}{0.9}

\usepackage{xcolor}
\definecolor{bittersweet}{rgb}{1.0, 0.44, 0.37}
\usepackage{subcaption}

\usepackage{pifont}% http://ctan.org/pkg/pifont
\usepackage{array}
\newcolumntype{P}[1]{>{\centering\arraybackslash}p{#1}}

\usepackage{microtype}

\newcommand{\ra}[1]{\renewcommand{\arraystretch}{#1}}
\begin{document}
% \renewcommand\thelinenumber{\color[rgb]{0.2,0.5,0.8}\normalfont\sffamily\scriptsize\arabic{linenumber}\color[rgb]{0,0,0}}
% \renewcommand\makeLineNumber {\hss\thelinenumber\ \hspace{6mm} \rlap{\hskip\textwidth\ \hspace{6.5mm}\thelinenumber}}
% \linenumbers
\pagestyle{headings}
\mainmatter
\def\ECCVSubNumber{100}  % Insert your submission number here

\title{TVR: A Large-Scale Dataset for Video-Subtitle Moment Retrieval} % Replace with your title

% INITIAL SUBMISSION 
\begin{comment}
\titlerunning{ECCV-20 submission ID \ECCVSubNumber} 
\authorrunning{ECCV-20 submission ID \ECCVSubNumber} 
\author{Anonymous ECCV submission}
\institute{Paper ID \ECCVSubNumber}
\end{comment}
%******************

% CAMERA READY SUBMISSION
%\begin{comment}
\titlerunning{TVR: A Large-Scale Dataset for Video-Subtitle Moment Retrieval}
% If the paper title is too long for the running head, you can set
% an abbreviated paper title here
%
\author{Jie Lei \and Licheng Yu \and Tamara L. Berg\index{Berg, Tamara L.} \and Mohit Bansal}
\authorrunning{Jie Lei, Licheng Yu, Tamara L. Berg\index{Berg, Tamara L.}, Mohit Bansal}
% First names are abbreviated in the running head.
% If there are more than two authors, 'et al.' is used.
%
\institute{University of North Carolina at Chapel Hill\\
\email{\{jielei, licheng, tlberg, mbansal\}@cs.unc.edu}}
%\end{comment}
%******************
\maketitle

\begin{abstract}
We introduce TV show Retrieval (\textbf{TVR}), a new multimodal retrieval dataset.
TVR requires systems to understand both videos and their associated subtitle (dialogue) texts, making it more realistic.
The dataset contains 109K queries collected on 21.8K videos from 6 TV shows of diverse genres, where each query is associated with a tight temporal window. 
The queries are also labeled with query types that indicate whether each of them is more related to video or subtitle or both, allowing for in-depth analysis of the dataset and the methods that built on top of it.
Strict qualification and post-annotation verification tests are applied to ensure the quality of the collected data. 
Further, we present several baselines and a novel Cross-modal Moment Localization (\textbf{XML}) network for multimodal moment retrieval tasks.
The proposed XML model uses a late fusion design with a novel Convolutional Start-End detector (\textbf{ConvSE}), surpassing baselines by a large margin and with better efficiency, providing a strong starting point for future work.
We have also collected additional descriptions for each annotated moment in TVR to form a new multimodal captioning dataset with 262K captions, named TV show Caption (\textbf{TVC}).\footnote{\footnotesize Published in ECCV 2020. Both datasets are publicly available. TVR: \url{https://tvr.cs.unc.edu}, TVC: \url{https://tvr.cs.unc.edu/tvc.html}.}
\end{abstract}

%%%%%%%%% BODY TEXT
\section{Introduction}\label{introuction}
\begin{figure}[!t]
  \centering
  \includegraphics[width=1.0\linewidth]{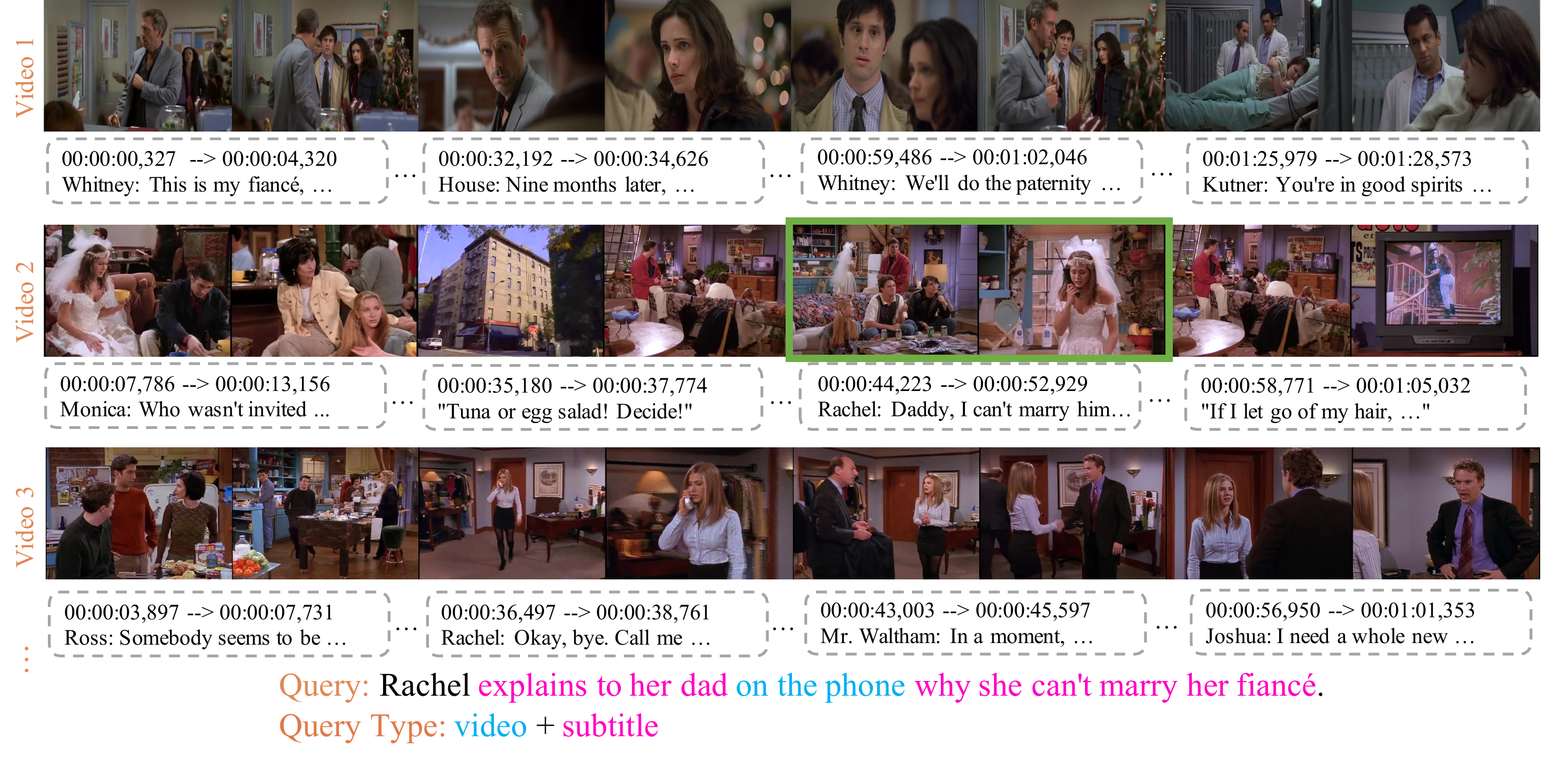}
  \caption{A TVR example in the VCMR task. Ground truth moment is shown in \textit{green box}. Colors in the query indicate whether the words are related to video (\textit{\textcolor{ cyan}{blue}}) or subtitle (\textit{\textcolor{magenta}{magenta}}) or both (\textit{black}). To better retrieve relevant moments from the video corpus, a system needs to comprehend both videos and subtitles}
  \label{fig:query_example}
\end{figure}

Enormous numbers of multimodal videos (with audio and/or text) are being uploaded to the web every day. To enable users to search through these videos and find relevant moments, an efficient and accurate method for retrieval of video data is crucial. 
Recent works~\cite{anne2017localizing,gao2017tall} introduced the task of Single Video Moment Retrieval (SVMR), whose goal is to retrieve a moment from a single video via a natural language query. 
Escorcia \emph{et al.}~\cite{escorcia2019temporal} extended SVMR to Video Corpus Moment Retrieval (VCMR), where a system is required to retrieve the most relevant moments from a large video corpus instead of from a single video. 
However, these works rely on a single modality (visual) as the context source for retrieval, as existing moment retrieval datasets~\cite{anne2017localizing,regneri2013grounding,gao2017tall,Krishna2017DenseCaptioningEI} are based on videos.
In practice, videos are often associated with other modalities such as audio or text, e.g., subtitles for movie/TV-shows or audience discourse accompanying live stream videos. 
These associated modalities could be equally important sources for retrieving user-relevant moments. Fig.~\ref{fig:query_example} shows a query example in the VCMR task, in which both videos and subtitles are vital to the retrieval process.

Hence, to study multimodal moment retrieval with both video and text contexts, we propose a new dataset - TV show Retrieval (\textbf{TVR}).
Inspired by recent works~\cite{tapaswi2016movieqa,kim2017deepstory,Lei2018TVQALC} that built multimodal datasets based on Movie/Cartoon/TV shows,
we select TV shows as our data resource as they typically involve rich social interactions between actors, involving both activities and dialogues.
During data collection, we present annotators with videos and associated subtitles to encourage them to write multimodal queries.
A tight temporal timestamp is labeled for each video-query pair. 
We do not use predefined fixed segments (as in~\cite{anne2017localizing}) but choose to freely annotate the timestamps for more accurate localization. 
Moreover, query types are collected for each query to indicate whether it is more related to the video, the subtitle, or both, allowing deeper analyses of systems.
To ensure data quality, we set up strict qualification and post-annotation quality verification tests.
In total, we have collected 108,965 high-quality queries on 21,793 videos from 6 TV shows, producing the largest dataset of this kind. 
Compared to existing datasets~\cite{anne2017localizing,regneri2013grounding,gao2017tall,Krishna2017DenseCaptioningEI}, we show TVR has greater linguistic diversity (Fig.~\ref{fig:query_diversity}) and involves more actions and people in its queries (Table~\ref{tab:dset_comparison_action_people}).

With the TVR dataset, we extend the moment retrieval task to a more realistic multimodal setup where both video and subtitle text need to be considered (i.e., `Video-Subtitle Moment Retrieval'). In this paper, we focus on the corpus-level task VCMR 
, as SVMR can be viewed as a simplified version of VCMR in which the ground-truth video is given beforehand. 
Prior works~\cite{anne2017localizing,gao2017tall,hendricks2018localizing,xu2019multilevel,ge2019mac,escorcia2019temporal} explore the moment retrieval task as a ranking problem over a predefined set of moment proposals. These proposals are usually generated using handcrafted heuristics~\cite{anne2017localizing,hendricks2018localizing} or sliding windows~\cite{gao2017tall,xu2019multilevel,ge2019mac,escorcia2019temporal} and are usually not temporally precise, leading to suboptimal performance. 
Furthermore, these methods may not be easily scaled to long videos: the number of proposals often increase quadratically with video length, making computational costs infeasible. 
Recent methods~\cite{ghosh2019excl,lei2019tvqa+} adapt start-end span predictors~\cite{seo2016bidirectional,chen2017reading} from the reading comprehension task to moment retrieval, by early fusion of video and language (query) features, then applying neural networks on the fused features to predict start-end probabilities. 
It has been shown~\cite{ghosh2019excl} that using span predictors outperforms several proposal-based methods. 
Additionally, start-end predictors allow a hassle-free extension to long videos, with only linearly increased computational cost. 
While~\cite{ghosh2019excl} has shown promising results in SVMR, it is not scalable to VCMR as it uses expensive early fusion operation. 
Consider retrieving $N$ queries in a corpus of $M$ videos, the approach in~\cite{ghosh2019excl} requires running several layers of LSTM~\cite{hochreiter1997long} on $M \mbox{$\cdot$}N$ early fused representations to generate the probabilities, which is computationally expensive for large values of $M$ and $N$. 

To address these challenges, we propose Cross-modal Moment Localization (\textbf{XML}), a late fusion approach for VCMR. In XML, videos (or subtitles) and queries are encoded independently, thus only $M\mbox{+}N$ neural network operations are needed. 
Furthermore, videos can be pre-encoded and stored. At test time, one only needs to encode new user queries, which greatly reduces user waiting time. Late fusion then integrates video and query representations with highly optimized matrix multiplication to generate 1D query-clip similarity scores over the temporal dimension of the videos. To produce moment predictions from these similarity scores, a naive approach is to rank the aforementioned sliding window proposals with confidence scores computed as the average of the similarity scores inside each proposal region. 
Alternatively, one can use TAG~\cite{zhao2017temporal} to progressively group top-scored clips. 
However, these methods rely on handcrafted rules and are not end-to-end trainable.
Inspired by image edge detectors~\cite{szeliski2010computer} in image processing, we propose Convolutional Start-End detector (\textbf{ConvSE}) that learns to detect start (up) and end (down) edges in the similarity signals with two trainable 1D convolution filters. 
Using the same backbone net, we show ConvSE has better performance than both approaches.
With late fusion and ConvSE, we further show XML outperforms previous methods~\cite{anne2017localizing,escorcia2019temporal,ghosh2019excl}, and does this with better computational efficiency.

To summarize, our contributions are three-fold: 
($i$) We introduce \textbf{TVR} dataset, a large-scale multimodal moment retrieval dataset with 109K high-quality queries of great linguistic diversity.
($ii$) We propose \textbf{XML}, an efficient approach that uses a late fusion design for the VCMR task. The core of XML is our novel \textbf{ConvSE} module which learns to detect start-end edges in 1D similarity signals with 2 convolution filters. Comprehensive experiments and analyses show XML surpasses all presented baselines by a large margin and runs with better efficiency. 
($iii$) We have also collected additional descriptions for each annotated moment in TVR to form a new multimodal captioning dataset with 262K captions, named TV show Caption (\textbf{TVC}).

\section{Related Work}\label{related_work}
The goal of natural language-based moment retrieval is to retrieve relevant moments from a single video~\cite{anne2017localizing,gao2017tall} or from a large video corpus~\cite{escorcia2019temporal}. 
In the following, we present a brief overview of the community efforts on these tasks and make distinctions between existing works and ours.

\kern1mm
\noindent\textbf{Datasets}.
Several datasets have been proposed for the task, e.g., DiDeMo~\cite{anne2017localizing}, ActivityNet Captions~\cite{Krishna2017DenseCaptioningEI}, CharadesSTA~\cite{gao2017tall}, and TACoS~\cite{regneri2013grounding}, where queries can be localized solely from video.
TVR differs from them by requiring additional text (subtitle) information in localizing the queries.
Two types of data annotation have been explored in previous works:
($i$) uniformly chunking videos into segments and letting an annotator pick one (or more) and write an unambiguous description~\cite{anne2017localizing}. 
For example, moments in DiDeMo~\cite{anne2017localizing} are created from fixed 5-second segments. 
However, such coarse temporal annotations are not well aligned with natural moments. 
In TVR, temporal windows are freely selected to more accurately capture important moments.
($ii$) converting a paragraph written for a whole video into separate query sentences~\cite{regneri2013grounding,gao2017tall,Krishna2017DenseCaptioningEI}.
While it is natural for people to use temporal connectives (e.g., `first', `then') and anaphora (e.g., pronouns)~\cite{rohrbach2014coherent} in a paragraph, these words make individual sentences less suitable as retrieval queries. 
In comparison, the TVR annotation process encourages annotators to write queries individually without requiring the context of a paragraph. 
Besides, TVR also has a larger size and greater linguistic diversity, see Sec.~\ref{subsec:data_analysis}.

\kern2mm
\noindent\textbf{Methods}.
Existing works~\cite{anne2017localizing,gao2017tall,hendricks2018localizing,xu2019multilevel,ge2019mac,escorcia2019temporal} pose moment retrieval as ranking a predefined set of moment proposals.
These proposals are typically generated with handcrafted rules~\cite{anne2017localizing,hendricks2018localizing} or sliding windows~\cite{gao2017tall,xu2019multilevel,ge2019mac,escorcia2019temporal}. 
Typically, such proposals are not temporally precise and are not scalable to long videos due to high computational cost.~\cite{gao2017tall,xu2019multilevel,ge2019mac} alleviate the first with a regression branch that offsets the proposals. However, they are still restricted by the coarseness of the initial proposals. 
Inspired by span predictors in reading comprehension~\cite{seo2016bidirectional,chen2017reading} and action localization~\cite{lin2018bsn}, we use start-end predictors to predict start-end probabilities from early fused query-video representations. 
Though these methods can be more flexibly applied to long videos and have shown promising performance on single video moment retrieval, the time cost of early fusion becomes unbearable when dealing with the corpus level moment retrieval problem: they require early fusing every possible query-video pair~\cite{escorcia2019temporal}.
Proposal based approaches MCN~\cite{anne2017localizing} and CAL~\cite{escorcia2019temporal} use a late fusion design, in which the video representations can be pre-computed and stored, making the retrieval more efficient. 
The final moment predictions are then made by ranking the Squared Euclidean Distances between the proposals w.r.t. a given query.
However, as they rely on predefined proposals, MCN and CAL still suffer from the aforementioned drawbacks, leading to less precise predictions and higher costs (especially for long videos).
Recent works~\cite{Zhang2018MANMA,chen2018temporally,zhang2019cross} consider word-level early fusion with the videos, which can be even more expensive.
In contrast, XML uses a late fusion design with a novel Convolutional Start-End (ConvSE) detector, which produces more accurate moment predictions while reducing the computational cost.

\section{Dataset}\label{sec:dataset}
Our TVR dataset is built on 21,793 videos from 6 long-running TV shows across 3 genres (\textit{sitcom}, \textit{medical}, \textit{crime}), provided by TVQA~\cite{Lei2018TVQALC}. 
Videos are paired with subtitles and are on average 76.2 seconds in length. 
In the following, we describe how we collected TVR and provide a detailed analysis of the data.

\subsection{Data Collection}\label{subsec:data_collection}
We used Amazon Mechanical Turk (AMT) for TVR data collection. 
Each AMT worker was asked to write a query using information from the video and/or subtitle, then mark the start and end timestamps to define a moment that matches the written query. 
This query-moment pair is required to be a unique match within the given video, i.e., the query should be a referring expression~\cite{kazemzadeh2014referitgame,anne2017localizing} that uniquely localizes the moment.
We additionally ask workers to select a query type from three types: \textit{video-only} - queries relevant to the visual content only, \textit{sub-only} - queries relevant to the subtitles only, and \textit{video+sub} - queries that involve both. 
In our pilot study, we found workers preferred to write \textit{sub-only} queries. 
A similar phenomenon was observed in TVQA~\cite{Lei2018TVQALC}, where people can achieve 72.88\% QA accuracy by reading the subtitles only. 
Therefore, to ensure that we collect a balance of queries requiring one or both modalities, we split the data annotation into two rounds - \textit{visual} round and \textit{textual} round. 
For the visual round, we encourage workers to write queries related to the visual content, including both \textit{video-only} and \textit{video+sub} queries. 
For the textual round, we encourage \textit{sub-only} and \textit{video+sub} queries. We ensure data quality with the following strategies:\footnote{We present a pipeline figure of our data collection procedure in Fig.~\ref{fig:tvr_collection_overview}.}

\noindent\textbf{Qualification Test}.
We designed a set of 12 multiple-choice questions as our qualification test and only let workers who correctly answer at least 9 questions participate in our annotation task, ensuring that workers understand our task requirements well.
In total, 1,055 workers participated in the test, with a pass rate of 67\%. 
Adding this qualification test greatly improved data quality.

\noindent\textbf{Automatic Check}. 
During collection, we used an automatic tool checking that all required annotations (query, timestamps, etc) have been performed and each query contains at least 8 words and is not copied from the subtitle. 

\noindent\textbf{Manual Check}.
Additional manual check of the collected data was done in house throughout the collection process. 
Those disqualified queries were re-annotated and workers with disqualified queries were removed from our worker list. 

\noindent\textbf{Post-Annotation Verification}. 
To verify the quality of the collected data, we performed a post-annotation verification experiment. 
We set up another AMT task where workers were required to rate the quality of the collected query-moment pairs based on \textit{relevance}, \textit{is the query-moment pair a unique-match}, etc. 
The rating was done in a \textit{likert-scale} manner with 5 options: \textit{strongly agree}, \textit{agree}, \textit{neutral}, \textit{disagree} and \textit{strongly disagree}. 
Results show that 92\% of the pairs have a rating of at least \textit{neutral}.
We further analyzed the group of queries that were rated as \textit{strongly disagree}, and found that 80\% of them were still of acceptable quality: e.g., slightly mismatched timestamps ($\leq$1 sec.). 
This verification was conducted on 3,600 query-moment pairs. 
Details are presented in Sec.~\ref{subsec:appendix_data_collection}. 

Given the high quality demonstrated by this verification, we did not further annotate each query, instead prioritizing collection toward  adding more TVR queries, and collecting additional captions for each annotated moment to form \textbf{TVC}, a large-scale multimodal video captioning dataset with 262K captions. See details in Sec.~\ref{sec:tvc}

\begin{table*}[t!]
\setlength{\tabcolsep}{0.15em}
\small
\centering
\caption{Comparison of TVR with existing moment retrieval datasets. \textit{Q} stands for query. \textit{Q context} indicate which modality the queries are related. \textit{Free st-ed} indicates whether the timestamps are freely annotated. \textit{Individual Q} means the queries are collected as individual sentences, rather than sentences in paragraphs}
\scalebox{0.64}{
\begin{tabular}{p{3.7cm}
P{1.4cm}
P{2.3cm}
P{1.3cm}P{1cm}P{2.2cm}P{0.9cm}P{0.9cm}P{0.9cm}P{1.2cm}P{1.8cm}}
\toprule
\multirow{2}{*}{\textbf{Dataset}} & \multirow{2}{*}{\textbf{Domain}} & \multirow{2}{*}{\textbf{\#Q/\#videos}} & \textbf{Vocab.}  & \textbf{Avg.} & \textbf{Avg. len. (s)}  & \multicolumn{2}{c}{\textbf{Q context}} & \textbf{Free} & \textbf{Q type} & \footnotesize{\textbf{Individual}} \\
 &  & & \textbf{size} & \textbf{Q len.} & moment/video & video & text & \textbf{st-ed} & \textbf{anno.} & \textbf{Q} \\ 
\midrule
TACoS~\cite{regneri2013grounding} & Cooking & 16.2K / 0.1K  & 2K & 10.5 & 5.9 / 287  & \checkmark & - & \checkmark & - & -\\ 
DiDeMo~\cite{anne2017localizing} & Flickr & 41.2K / 10.6K & 7.6K & 8.0 & 6.5 / 29.3  & \checkmark & - & - & - & \checkmark\\
ActivityNet Captions~\cite{Krishna2017DenseCaptioningEI} & Activity & 72K / 15K & 12.5K & 14.8 & 36.2 / 117.6  & \checkmark & - & \checkmark & - & -\\
CharadesSTA~\cite{gao2017tall} & Activity & 16.1K / 6.7K  & 1.3K & 7.2 & 8.1 / 30.6 & \checkmark & - & \checkmark & - & -\\
\midrule
TVR & TV show & 109K / 21.8K  & 57.1K & 13.4 & 9.1 / 76.2 & \checkmark & \checkmark & \checkmark & \checkmark & \checkmark \\
\bottomrule
\end{tabular}
}
\label{tab:dset_comparison}
\end{table*}

\begin{figure*}[!t]
  \centering
   \includegraphics[width=0.98\textwidth]{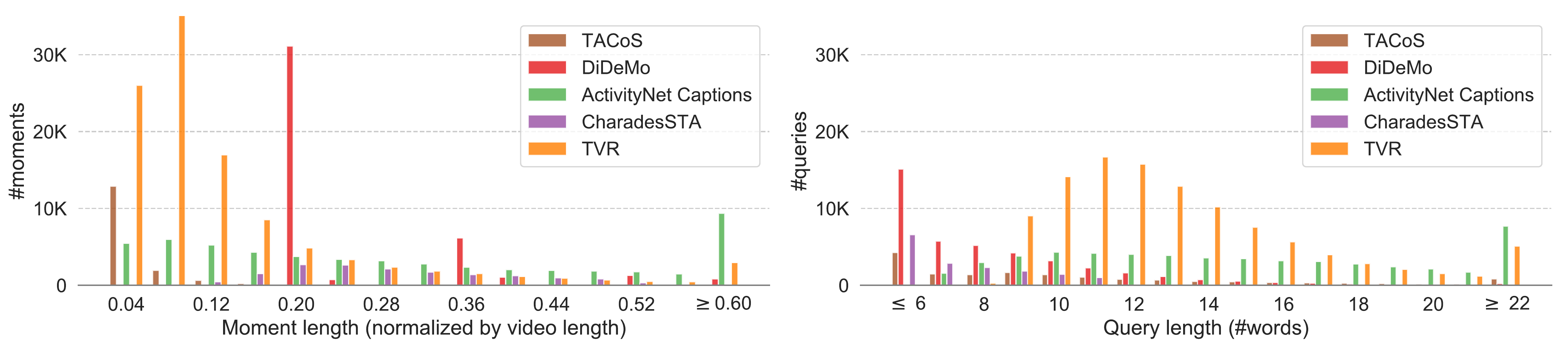}
  \caption{Distributions of moment (\textit{left}) and query (\textit{right}) lengths. Compared to existing moment retrieval datasets~\cite{regneri2013grounding,anne2017localizing,Krishna2017DenseCaptioningEI,gao2017tall}, TVR has relatively shorter moments (normalized) and longer queries. Best viewed digitally with zoom}
  \label{fig:length_dist} 
\end{figure*}

\begin{figure*}[!t]
  \centering
  \includegraphics[width=0.98\textwidth]{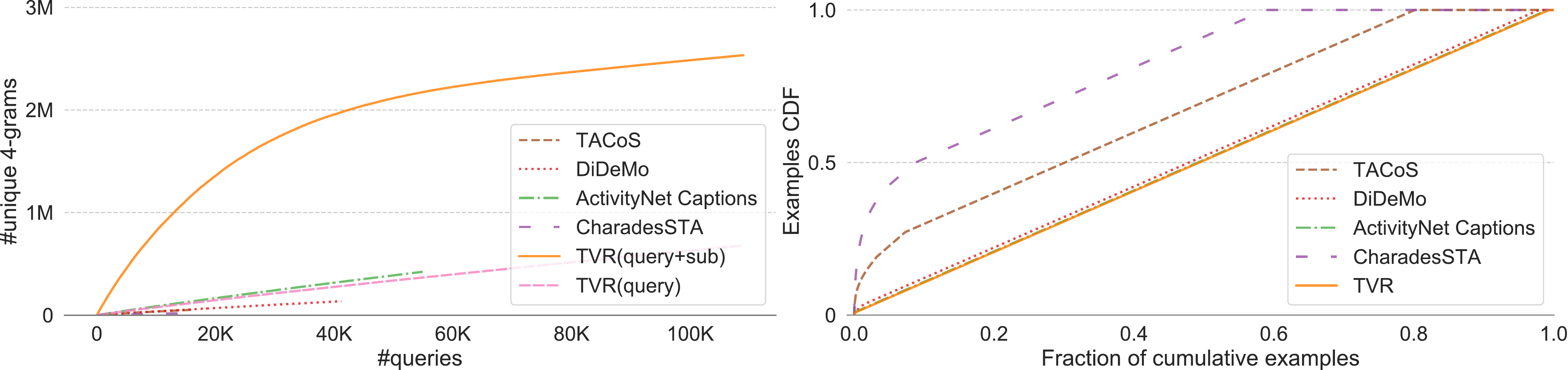}
  \caption{\textit{Left}: \#unique 4-gram as a function of \#queries. \textit{Right}: CDF of queries ordered by frequency, to obtain this plot, we sampled 10K queries from each dataset, we consider two queries to be the same if they exact match, after tokenization and lemmatization, following~\cite{zellers2019recognition}. Compared to existing moment retrieval datasets~\cite{regneri2013grounding,anne2017localizing,Krishna2017DenseCaptioningEI,gao2017tall}, TVR has greater diversity, i.e., it has more unique 4-grams and almost every TVR query is unique. Best viewed digitally with zoom}
  \label{fig:query_diversity}
\end{figure*}

\begin{table}[t]
\setlength{\tabcolsep}{0.05em}
\centering
\small
\caption{Percentage of queries that have multiple actions or involve multiple people. Statistics is based on 100 manually labeled queries from each dataset. We also show query examples, with unique person mentions \underline{underlined} and actions in \textbf{bold}. Compared to existing datasets, TVR queries typically have more people and actions and require both \textit{video} and \textit{sub} (subtitle) context
} 
\scalebox{0.66}{
\begin{tabular}{p{2.2cm}P{1.5cm}P{1.5cm}c}
\toprule
\multirow{2}{*}{\textbf{Dataset}} & \textbf{\#actions} & \textbf{\#people} & \multirow{2}{*}{\textbf{Query examples} (\textit{query type})} \\
 & \textbf{$\geq$2 (\%)} & \textbf{$\geq$2 (\%)} &  \\
\midrule
\multirow{2}{*}{TACoS~\cite{regneri2013grounding}} & \multirow{2}{*}{20} & \multirow{2}{*}{0} & \multicolumn{1}{l}{\underline{She} \textbf{rinses} the peeled carrots off in the sink. (\textit{video})}\\
& & & \multicolumn{1}{l}{The \underline{person} \textbf{removes} roots and outer leaves and \textbf{rewashes} the leek. (\textit{video})} \\
\midrule
\multirow{2}{*}{CharadesSTA~\cite{gao2017tall}} & \multirow{2}{*}{6} & \multirow{2}{*}{12} & \multicolumn{1}{l}{A \underline{person} is \textbf{eating} food slowly. (\textit{video})}\\
& & & \multicolumn{1}{l}{A \underline{person} is \textbf{opening} the door to a bedroom. (\textit{video})}\\
\midrule
ActivityNet  & \multirow{2}{*}{44} & \multirow{2}{*}{44} & \multicolumn{1}{l}{\underline{He} then \textbf{grabs} a metal mask and \textbf{positions} himself correctly on the floor. (\textit{video})}\\
Caption~\cite{Krishna2017DenseCaptioningEI} & & & \multicolumn{1}{l}{The same \underline{man} \textbf{comes} back and \textbf{lifts} the weight over his head again. (\textit{video})}\\
\midrule
\multirow{2}{*}{DiDeMo~\cite{anne2017localizing}} & \multirow{2}{*}{6} & \multirow{2}{*}{10} & \multicolumn{1}{l}{A dog \textbf{shakes} its body. (\textit{video})}\\
& & & \multicolumn{1}{l}{A \underline{lady} in a cowboy hat \textbf{claps} and \textbf{jumps} excitedly. (\textit{video})}\\
\midrule
\multirow{3}{*}{TVR} & \multirow{3}{*}{67} & \multirow{3}{*}{66} & \multicolumn{1}{l}{\underline{Bert} \textbf{leans} down and \textbf{gives} \underline{Amy} a hug who is \textbf{standing} next to \underline{Penny}. (\textit{video})} \\
& & & \multicolumn{1}{l}{\underline{Taub} \textbf{argues} with the \underline{patient} that fighting in Hockey \textbf{undermines} the sport. (\textit{sub})} \\
& & & \multicolumn{1}{l}{\underline{Chandler} \textbf{points} at \underline{Joey} while \textbf{describing} a \underline{woman} who wants to \textbf{date} him.  (\textit{video+sub})} \\
\bottomrule
\end{tabular}
}
\label{tab:dset_comparison_action_people}
\end{table}

\subsection{Data Analysis and Comparison}\label{subsec:data_analysis}
Table~\ref{tab:dset_comparison} shows an overview of TVR and its comparisons with existing moment retrieval datasets~\cite{regneri2013grounding,gao2017tall,Krishna2017DenseCaptioningEI,anne2017localizing}.
TVR contains 109K human annotated query-moment pairs on 21.8K videos, making it the largest of its kind. 
Moments have an average length of 9.1 seconds, and are annotated with tight start and end timestamps, enabling training and evaluating on more precise localization. 
Compared to existing datasets, TVR has relatively shorter (video-length normalized) moments and longer queries (Fig.~\ref{fig:length_dist}). 
It also has greater linguistic diversity (Fig.~\ref{fig:query_diversity}): it has more unique 4-grams and almost every query is unique, making the textual understanding of TVR more challenging. 
As TVR is collected on TV shows, query-moment matching often involves understanding rich interactions between characters. 
Table~\ref{tab:dset_comparison_action_people} shows a comparison of the percentages of queries that involve more than one action or person across different datasets. 
66\% of TVR queries involve at least two people and 67\% involve at least two actions, both of which are significantly higher than those of other datasets. 
This makes TVR an interesting testbed for studying multimodal interactions between people.
Additionally, each TVR query is labeled with a query type, indicating whether this query is based on video, subtitle or both, which can be used for deeper analyses of the systems.

\section{Cross-modal Moment Localization (XML)}\label{sec:methods}

In VCMR, the goal is to retrieve a moment from a large video corpus $V\mbox{=} \{v_i\}_{i=1}^{n}$ given a query $q_j$. 
Each video $v_i$ is represented as a list of consecutive short clips, i.e., $v_i \mbox{=} [c_{i, 1}, c_{i, 2}, ..., c_{i, l}]$. In TVR, each short clip is also associated with temporally aligned subtitle sentences.
The retrieved moment is denoted as $v_i[t_{st}\text{:}t_{ed}] \mbox{=} [c_{i, t_{st}}, c_{i, t_{st}+1}, ..., c_{i, t_{ed}}]$.
To address VCMR, we propose a hierarchical Cross-modal Moment Localization (XML) network. 
XML performs video retrieval (VR) in its shallower layers and more fine-grained moment retrieval in its deeper layers.
It uses a late fusion design with a novel Convolutional Start-End (ConvSE) detector, making the moment predictions efficient and accurate.

\begin{figure*}[!t]
  \centering
  \includegraphics[width=0.96\linewidth]{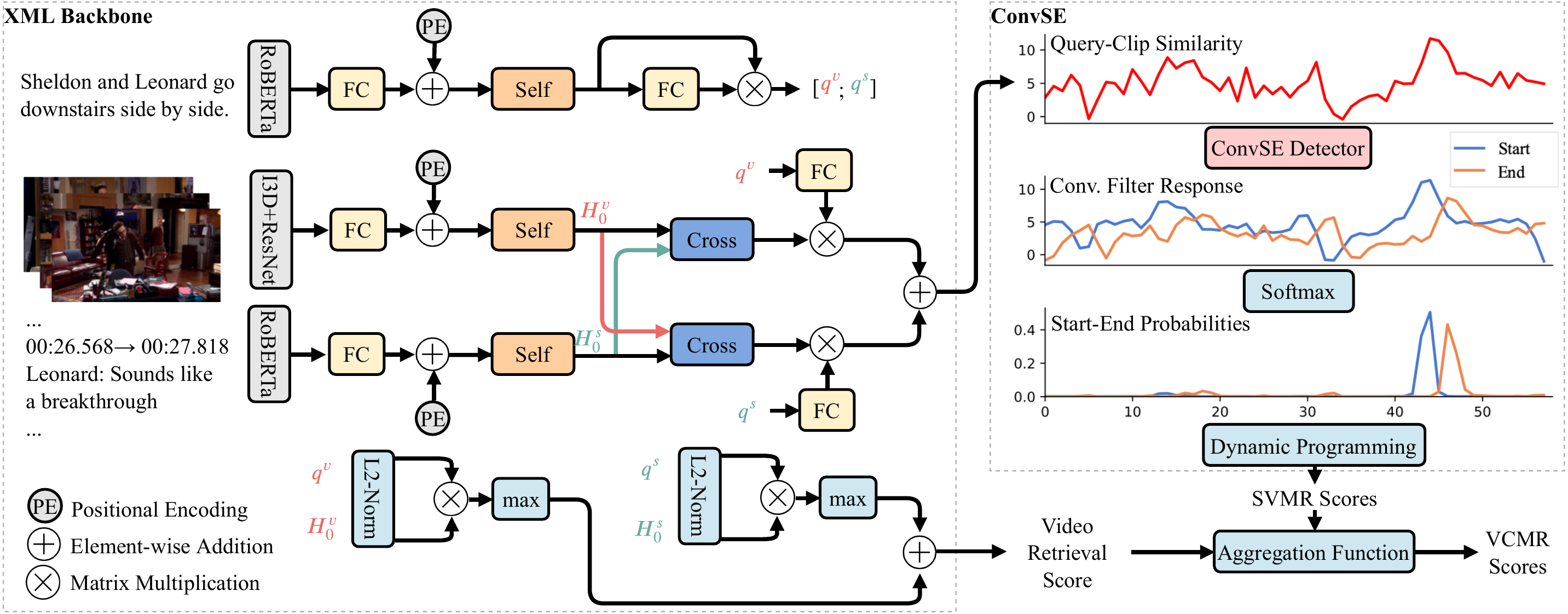}
  \caption{Cross-modal Moment Localization (XML) model overview. \textit{Self}=\textit{Self Encoder}, \textit{Cross}=\textit{Cross Encoder}.  We describe \textit{XML Backbone} in Sec.~\ref{subsec:xml_backbone}, \textit{ConvSE} module in Sec.~\ref{subsec:convse} and show XML's training and inference procedure in Sec.~\ref{subsec:training_inference}}
  \label{fig:model_overview}
\end{figure*}

\subsection{XML Backbone Network}\label{subsec:xml_backbone}
\noindent\textbf{Input Representations}. To represent videos, we consider both appearance and motion features. For appearance, we extract 2048D ResNet-152~\cite{he2016deep} features at 3FPS and max-pool the features every 1.5 seconds to get a clip-level feature. 
For motion, we extract 1024D I3D~\cite{carreira2017quo} features every 1.5 seconds. 
The ResNet-152 model is pre-trained on ImageNet~\cite{deng2009imagenet} for image recognition, and the I3D model is pre-trained on Kinetics-600~\cite{kay2017kinetics} for action recognition. 
The final video representation is the concatenation of the two features after L2-normalization, denoted as $E^{v} \in \mathbb{R}^{l \times 3072}$, where $l$ is video length (\#clips).
We extract contextualized text features using a 12-layer RoBERTa~\cite{liu2019roberta}. 
Specifically, we first fine-tune RoBERTa using the queries and subtitle sentences in TVR train-split with MLM objective~\cite{devlin2018bert}, then fix the parameters to extract contextualized token embeddings from its second-to-last layer~\cite{lei2019tvqa+}. 
For queries, we directly use the extracted token embeddings, denoted as $E^{q} \in \mathbb{R}^{l_q \times 768}$, where $l_q$ is query length (\#words). 
For subtitles, we first extract token-level embeddings, then max-pool them every 1.5 seconds to get a 768D clip-level feature vector. 
We use a 768D zero vector if encountering no subtitle.
The final subtitle embedding is denoted as $E^{s} \in \mathbb{R}^{l \times 768}$.
The extracted features are projected into a low-dimensional space via a linear layer with ReLU~\cite{glorot2011deep}. 
We then add learned positional encoding~\cite{devlin2018bert} to the projected features. 
Without ambiguity, we reuse the symbols by denoting the processed features as $E^{v} \in \mathbb{R}^{l \times d}$, $E^{s} \in \mathbb{R}^{l \times d}$, $E^{q} \in \mathbb{R}^{l_q \times d}$, where $d$ is hidden size.

\kern1mm
\noindent\textbf{Query Encoding}.
As TVR queries can be related to either video or subtitle, we adopt a modular design to dynamically decompose the query into two modularized vectors.
Specifically, the query feature is encoded using a \textit{Self-Encoder}, consisting of a self-attention~\cite{vaswani2017attention} layer and a linear layer, with a residual~\cite{he2016deep} connection followed by layer normalization~\cite{ba2016layer}. 
We denote the encoded query as $H^{q} \in \mathbb{R}^{l_q \times d}$.
Then, we apply two trainable modular weight vectors $\mathbf{w}_{m} \in \mathbb{R}^{d}$, $m \in \{v, s\}$ to compute the attention scores of each query word w.r.t. the video ($v$) or subtitle ($s$). 
The scores are used to aggregate the information of $H^{q}\mbox{=}\{\mathbf{h^{q}_{r}}\}_{r=1}^{l_q}$ to generate modularized query vectors $\mathbf{q}^{m} \in \mathbb{R}^{d}$~\cite{yu2018mattnet}:
\begin{align}
    a^{m}_r = \frac{\mathrm{exp}(\mathbf{w}_m^T \mathbf{h}^{q}_r)}{\sum_{k=1}^{l_q} \mathrm{exp}(\mathbf{w}_m^T \mathbf{h}^{q}_k)}, \;\;\mathbf{q}^m = \sum_{r=1}^{l_q} a^{m}_r \mathbf{h^{q}_{r}}, \;\;\mathrm{where}\;m \in \{v, s\}.
\end{align}

\kern1mm
\noindent\textbf{Context Encoding}.
Given the video and subtitle features $E^v$, $E^s$, we use two Self-Encoders to compute their single-modal contextualized features $H^{v}_{0} \in \mathbb{R}^{l \times d}$ and $H^{s}_{0} \in \mathbb{R}^{l \times d}$. 
Then, we encode their cross-modal representations via \textit{Cross-Encoder}.
which takes as input the self-modality and cross-modality features, and encodes the two via  cross-attention~\cite{vaswani2017attention} followed by a linear layer, a residual connection, a layer normalization, and another Self-Encoder.
We denote the final video and subtitle representations as $H^{v}_{1} \in \mathbb{R}^{l \times d}$ and $H^{s}_{1} \in \mathbb{R}^{l \times d}$, respectively.

\subsection{Convolutional Start-End Detector}\label{subsec:convse}
Given $H^{v}_{1}, H^{s}_{1}$ and $\mathbf{q}^v, \mathbf{q}^s$, we compute query-clip similarity scores $S_{\mathrm{query\mbox{-}clip}} \in \mathbb{R}^{l}$:
\begin{align}
    \label{eq:query-clip}
    S_{\mathrm{query\mbox{-}clip}} = \frac{1}{2}(H^{v}_{1} \mathbf{q}^v +  H^{s}_{1} \mathbf{q}^s).
\end{align}

To produce moment predictions from $S_{\mathrm{query\mbox{-}clip}}$, one could rank sliding window proposals with confidence scores computed as the average of scores in each proposal region, or use TAG~\cite{zhao2017temporal} to progressively group top-scored regions. However, both methods require handcrafted rules and are not trainable. Inspired by edge detectors in image processing~\cite{szeliski2010computer}, we propose Convolutional Start-End detector (\textbf{ConvSE}) with two 1D convolution filters to learn to detect start (up) and end (down) edges in the score curves. Clips inside a semantically close span will have higher similarity to the query than those outside, naturally forming detectable edges around the span. Fig.~\ref{fig:model_overview} (right) and Fig.~\ref{fig:conv_example} show examples of the learned ConvSE filters applied to the similarity curves. Specifically, we use two trainable filters (no bias) to generate the start ($\mathrm{st}$) and end ($\mathrm{ed}$) scores:
\begin{align}
    S_{\mathrm{st}} = \mathrm{Conv1D}_{\mathrm{st}}(S_{\mathrm{query\mbox{-}clip}}), \;\;
    S_{\mathrm{ed}} = \mathrm{Conv1D}_{\mathrm{ed}}(S_{\mathrm{query\mbox{-}clip}}).
\end{align}
The scores are normalized with softmax to output the probabilities $P_{\mathrm{st}}, P_{\mathrm{ed}} \in \mathbb{R}^{l}$. 
In Sec.~\ref{subsec:model_analysis}, we show ConvSE outperforms the baselines and is also interpretable.

\subsection{Training and Inference}\label{subsec:training_inference}
\noindent\textbf{Video Retrieval}.
Given the modularized queries $\mathbf{q}^{v}, \mathbf{q}^{s}$ and the encoded contexts $H^{v}_{0}, H^{s}_{0}$, we compute the video-level retrieval (VR) score as:
\begin{align}
    s^{\mathrm{vr}} = \frac{1}{2}\sum_{m\in \{v, s\}} \mathrm{max}(\, \frac{H^{m}_{0}}{\left\lVert H^{m}_{0}\right\rVert} \frac{\mathbf{q}^{m}}{\left\lVert \mathbf{q}^{m}\right\rVert}\, ).
\end{align}
This essentially computes the cosine similarity between each clip and query and picks the maximum.
The final VR score is the average of the scores from the two modalities.
During training, we sample two negative pairs $(q_i, v_j)$ and $(q_z, v_i)$ for each positive pair of $(q_i, v_i)$ to calculate a combined hinge loss as~\cite{yu2018mattnet}:
\begin{align}
    L^{\mathrm{vr}} =& \frac{1}{n} \sum_i [\mathrm{max}(0, \Delta + s^{\mathrm{vr}}(v_j|q_i) - s^{\mathrm{vr}}(v_i|q_i)) \nonumber \\
    &+ \mathrm{max}(0, \Delta + s^{\mathrm{vr}}(v_i|q_z) - s^{\mathrm{vr}}(v_i|q_i)) ].
\end{align}

\kern1mm
\noindent\textbf{Single Video Moment Retrieval}.
Given the start, end probabilities $P_{\mathrm{st}}, P_{\mathrm{ed}}$, we define single video moment retrieval loss as:
\begin{align}
\label{eq:svmr_score}
L^{\mathrm{svmr}} = - \frac{1}{n} \sum_i [\mathrm{log}(P_{i, \mathrm{st}}(t_{\mathrm{st}}^i)) + \mathrm{log}(P_{i, \mathrm{ed}}(t_{\mathrm{ed}}^i))],
\end{align}
where $t_{\mathrm{st}}^i$ and $t_{\mathrm{ed}}^i$ are the ground-truth indices.
At inference, predictions can be generated from the probabilities in linear time using dynamic programming~\cite{seo2016bidirectional}. The confidence score of a predicted moment $[ t_{\mathrm{st}}^{'}, t_{\mathrm{ed}}^{'}]$ is computed as: 

\begin{align}
s^{\mathrm{svmr}}(t_{\mathrm{st}}^{'}, t_{\mathrm{ed}}^{'}) = P_{\mathrm{st}}( t_{\mathrm{st}}^{'})P_{\mathrm{ed}}( t_{\mathrm{ed}}^{'}), \;t_{\mathrm{st}}^{'} \leq  t_{\mathrm{ed}}^{'}.
\end{align}
\noindent
To use length prior, we add an additional constraint $L_{min} \leq t_{\mathrm{ed}}^{'} - t_{\mathrm{st}}^{'} + 1 \leq L_{max}$. For TVR, we set $L_{min}\mbox{=}2$ and $L_{max}\mbox{=}16$ for clip length 1.5 seconds.

\kern1mm
\noindent\textbf{Video Corpus Moment Retrieval}. 
Our final training loss combines both: $L^{\mathrm{vcmr}} = L^{\mathrm{vr}} + \lambda L^{\mathrm{svmr}}$, where the hyperparameter $\lambda$ is set as 0.01.
At inference, we compute the VCMR score with the following aggregation function:
\begin{align}
\label{eq:vcmr_score}
s^{\mathrm{vcmr}}(v_j, t_{\mathrm{st}}, t_{\mathrm{ed}} | q_i) =  s^{\mathrm{svmr}}(t_{\mathrm{st}}, t_{\mathrm{ed}} | v_j, q_i) \mathrm{exp}(\alpha s^{\mathrm{vr}}(v_j | q_i)),
\end{align}
where $s^{\mathrm{vcmr}}(v_j, t_{\mathrm{st}}, t_{\mathrm{ed}} | q_i)$ is the retrieval score of moment $v_j[t_{\mathrm{st}}\mbox{:} t_{\mathrm{ed}}]$ w.r.t. the query $q_i$. 
The exponential term and the hyperparameter $\alpha$ are used to balance the importance of the two scores. 
A higher $\alpha$ encourages more moments from top retrieved videos. 
Empirically, we find $\alpha\mbox{=}20$ works well. 
At inference, for each query, we first retrieve the top 100 videos based on $s^{\mathrm{vr}}$, then rank all the moments in the 100 videos by $s^{\mathrm{vcmr}}$ to give the final predictions. 

\section{Experiments}\label{experiments}

\subsection{Data, Metrics and Implementation Details}
\noindent\textbf{Data}.
TVR contains 109K queries from 21.8K videos. 
We split TVR into 80\% \textit{train}, 10\% \textit{val}, 5\% \textit{test-public} and 5\% \textit{test-private} splits such that videos and their associated queries appear in only one split. 
\textit{test-public} will be used for a public leaderboard, \textit{test-private} is reserved for future challenges. 

\kern1mm
\noindent\textbf{Metrics}.
Following~\cite{escorcia2019temporal,gao2017tall}, we use average recall at K (R@K) over all queries as our metric. 
A prediction is correct if: ($i$) predicted video matches the ground truth; ($ii$) predicted span has high overlap with the ground truth where 
temporal intersection over union (IoU) is used to measure overlap. 

\kern1mm
\noindent\textbf{Implementation Details}.
All baseline comparisons are configured to use the same hidden size as XML. We train the baselines following the original papers.
We use the same features for all the models. 
To support retrieval using subtitle for the baselines, we add a separate subtitle stream and average the final predictions from both streams.
Non-maximum suppression is not used as we do not observe consistent performance gain on the \textit{val} set.

\begin{table*}[!t]
\setlength{\tabcolsep}{0.3em}
\ra{1.}
\centering
\small
\caption{Baseline comparison on TVR \textit{test-public} set, VCMR task. Model references: \textit{MCN}~\cite{anne2017localizing},
\textit{CAL}~\cite{escorcia2019temporal}, \textit{MEE}~\cite{miech2018learning}, \textit{ExCL}~\cite{ghosh2019excl}. Results with TEF~\cite{anne2017localizing} feature are presented in Table~\ref{tab:main_vcmr_res_tef}}
\scalebox{0.75}{
\begin{tabular}{lP{1.4cm}P{1.4cm}
>{\raggedleft\arraybackslash}p{0.9cm}
>{\raggedleft\arraybackslash}p{0.9cm}
>{\raggedleft\arraybackslash}p{0.9cm}
>{\raggedleft\arraybackslash}p{0.9cm}
>{\raggedleft\arraybackslash}p{0.9cm}
>{\raggedleft\arraybackslash}p{0.9cm}
>{\raggedleft\arraybackslash}p{0.9cm}
>{\raggedleft\arraybackslash}p{0.9cm}
r
}
\toprule
\multirow{2}{*}{Model} & \multirow{2}{*}{w/ video} & \multirow{2}{*}{w/ sub.} & \multicolumn{4}{c}{IoU=0.5} & \multicolumn{4}{c}{IoU=0.7} & \multicolumn{1}{c}{Runtime~$\downarrow$} \\  
\cmidrule(l){4-7}  \cmidrule(l){8-11}
 & & & R@1 & R@5 & R@10 & R@100 & R@1 & R@5 & R@10 & R@100 & \multicolumn{1}{c}{(seconds)}\\ 
\midrule
Chance & - & - & 0.00 & 0.02 & 0.04 & 0.33 & 0.00 & 0.00 & 0.00 & 0.07 \\
\multicolumn{3}{l}{\textbf{Proposal based Methods}} & & & & &  &  &  & & \\
MCN & \checkmark & \checkmark & 0.02 & 0.15 & 0.24 & 2.20 & 0.00 & 0.07 & 0.09 & 1.03 & -\\
CAL & \checkmark & \checkmark & 0.09 & 0.31 & 0.57 & 3.42 & 0.04 & 0.15 & 0.26 & 1.89 & - \\
\multicolumn{3}{l}{\textbf{Retrieval + Re-ranking}} & & & &  &  &  & & \\
MEE+MCN & \checkmark & \checkmark & 0.92 & 3.69 & 5.58 & 17.91 & 0.42 & 1.89 & 2.98 & 10.84 & 66.8 \\
MEE+CAL & \checkmark & \checkmark & 0.97 & 3.75 & 5.80 & 18.66 & 0.39 & 1.69 & 2.98 & 11.52 & 161.5 \\ 
MEE+ExCL & \checkmark & \checkmark & 0.92 & 2.53 & 3.60 & 6.01 & 0.33 & 1.19 & 1.73 & 2.87 & 1307.2 \\
\midrule
XML & \checkmark & \checkmark & \textbf{7.25} & \textbf{16.24} & \textbf{21.65} & \textbf{44.44} & \textbf{3.25} & \textbf{8.71} & \textbf{12.49} & \textbf{29.51} & \textbf{25.5} \\
\bottomrule
\end{tabular}
}
\label{tab:main_vcmr_res}
\end{table*}

\subsection{Baselines Comparison}\label{subsec:baseline_comparison}
In this section, we compare XML with baselines on TVR \textit{test-public} set (5,445 queries and 1,089 videos). We report the runtime for top-performing methods, averaged across 3 runs on an RTX 2080Ti GPU. Time spent on data loading, pre-processing, backend model (i.e., ResNet-152, I3D, RoBERTa) feature extraction, etc, is ignored since they should be similar for all methods. We mainly focus on the VCMR task here. In Sec.~\ref{subsec:additional_tvr_experiments} and Sec.~\ref{didemo_experiments}, we include additional experiments: (1) model performance on single video moment retrieval and video retrieval tasks; (2) computation and storage cost comparison in a 1M videos corpus; (3) Temporal Endpoint Feature (TEF)~\cite{anne2017localizing} model results; (4) feature and model architecture ablation studies; (5) VCMR results on DiDeMo~\cite{anne2017localizing} dataset, etc.

\kern1mm
\noindent\textbf{Proposal based Methods}. MCN~\cite{anne2017localizing} and CAL~\cite{escorcia2019temporal} pose the moment retrieval task as a ranking problem in which all moment proposal candidates are ranked based on their squared Euclidean Distance with the queries. 
For VCMR, they require directly ranking all the proposals (95K in the following experiments) in the video corpus for each query, which can be costly and difficulty.
In contrast, XML uses a hierarchical design that performs video retrieval in its shallow layers and moment retrieval on the retrieved videos in its deeper layers. 
In Table~\ref{tab:main_vcmr_res}, XML is showing to have significantly higher performance than MCN and CAL.

\kern1mm
\noindent\textbf{Retrieval+Re-ranking Methods}.
We also compare to methods under the retrieval+re-ranking setting~\cite{escorcia2019temporal} where we first retrieve a set of candidate videos using a given method and then re-rank the moment predictions in the candidate videos using another method. 
Specifically, we first use MEE~\cite{miech2018learning} to retrieve 100 videos for each query as candidates. 
Then, we use MCN and CAL to rank all of the proposals in the candidate videos. ExCL~\cite{ghosh2019excl} is an early fusion method designed for SVMR, with a start-end predictor. 
We adapt it to VCMR by combining MEE video-level scores with ExCL moment-level scores, using Eq.~\ref{eq:vcmr_score}. 
The results are shown in Table~\ref{tab:main_vcmr_res}. 
Compared to their purely proposal based counterparts (i.e., MCN and CAL), both MEE+MCN and MEE+CAL achieve significant performance gain, showing the benefit of reducing the number of proposals needed to rank (by reducing the number of videos). 
However, they are still far below XML as they use very coarse-grained, predefined proposals.
In Sec.~\ref{subsec:model_analysis}, we show our start-end detector performs consistently better than predefined proposals~\cite{escorcia2019temporal,zhao2017temporal} under our XML framework.
Compared to MEE+ExCL, XML achieves $9.85\times$ performance gain (3.25 \textit{vs.} 0.33, R@1 IoU=0.7) and $51.3\times$ speedup (25.5s \textit{vs.} 1307.2s). 
In the Sec.~\ref{subsec:more_vcmr_cpmparison}, we show that this speedup can be even more significant (287$\times$) when retrieving on a larger scale video corpus (1M videos) with pre-encoded video representations.
This huge speedup shows the effectiveness of XML's late fusion design over ExCL's early fusion design.

\begin{figure*}[!t]
  \centering
  \includegraphics[width=0.96\linewidth]{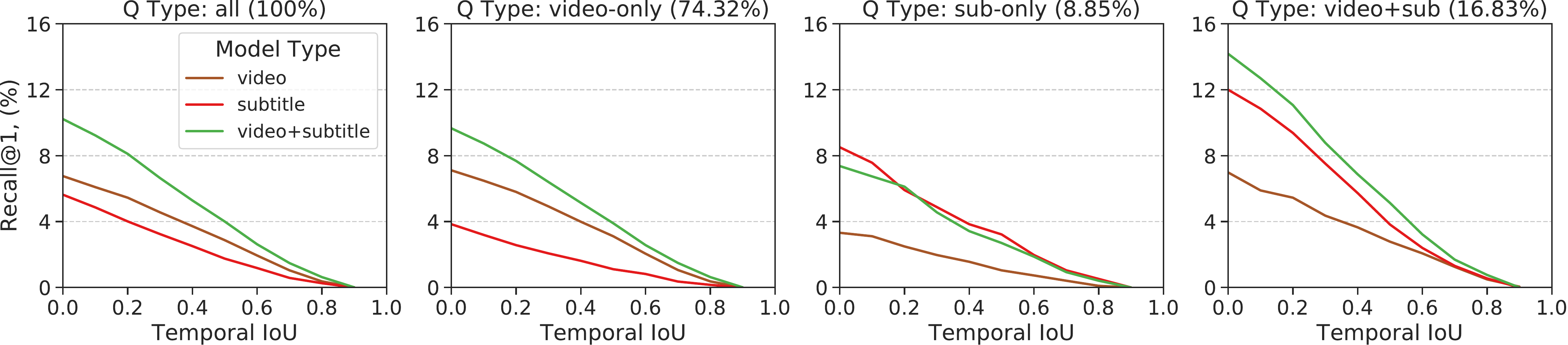}
  \caption{Performance breakdown of XML models that use only \textit{video}, \textit{subtitle}, or both as inputs, by different query types (with percentage of queries shown in brackets). The performance is evaluated on TVR \textit{val} set for VCMR}
  \label{fig:qtype_performance}
\end{figure*}

\subsection{Model Analysis}\label{subsec:model_analysis}

\noindent\textbf{Video vs. Subtitle}.
In Fig.~\ref{fig:qtype_performance}, we compare to XML variants that use only video or subtitle. 
We observe that the full \textit{video+subtitle} model has better overall performance than single modality models (\textit{video} and \textit{subtitle}), demonstrating that both modalities are useful. 
We also see that a model trained on one modality does not perform well on the queries tagged by another modality, e.g., the \textit{video} model performs much worse on \textit{sub-only} queries compared to the \textit{subtitle} model. 

\begin{figure}[!t]
  \centering
  \includegraphics[width=0.96\linewidth]{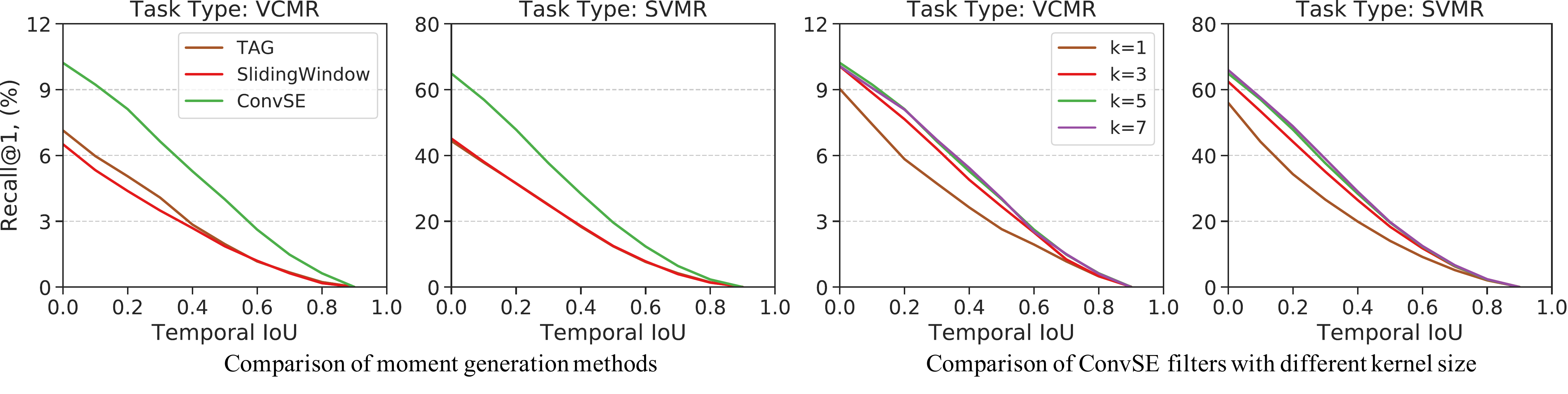}
  \caption{ConvSE Analysis. \textit{Left}: comparison of moment generation methods. \textit{Right}: comparison of ConvSE filters with different kernel sizes (\textit{k})}
  \label{fig:convse_comparison}
\end{figure}

\kern1mm
\noindent\textbf{ConvSE: Comparison and Analysis}. 
To produce moment predictions from the query-clip similarity signals, we proposed ConvSE that learns to detect start (up) and end (down) edges in the 1D similarity signals. 
To show its effectiveness, we compare ConvSE with two baselines under our XML backbone network: (1) sliding window, where we rank proposals generated by multi-scale sliding windows, with proposal confidence scores calculated as the average of scores inside each proposal region. On average, it produces 87 proposals per video. The proposals used here are the same as the ones used for MCN and CAL in our previous experiments; (2) TAG~\cite{zhao2017temporal} that progressively groups top-scored clips with the classical watershed algorithm~\cite{roerdink2000watershed}. 
Since these two methods do not produce start-end probabilities, we cannot train the model with the objective in Eq.~\ref{eq:svmr_score}. 
Thus, we directly optimize the query-clip similarity scores in Eq.\ref{eq:query-clip} with Binary Cross Entropy loss: we assign a label of 1 if the clip falls into the ground-truth region, 0 otherwise.
While both sliding window and TAG approaches rely on handcrafted rules, ConvSE \textbf{learns from data}. 
We show in Fig.~\ref{fig:convse_comparison} (left), under the same XML backbone network, ConvSE has consistent better performance across all IoU thresholds on both VCMR and SVMR tasks.

\begin{figure}[!t]
  \centering
  \includegraphics[width=0.98\linewidth]{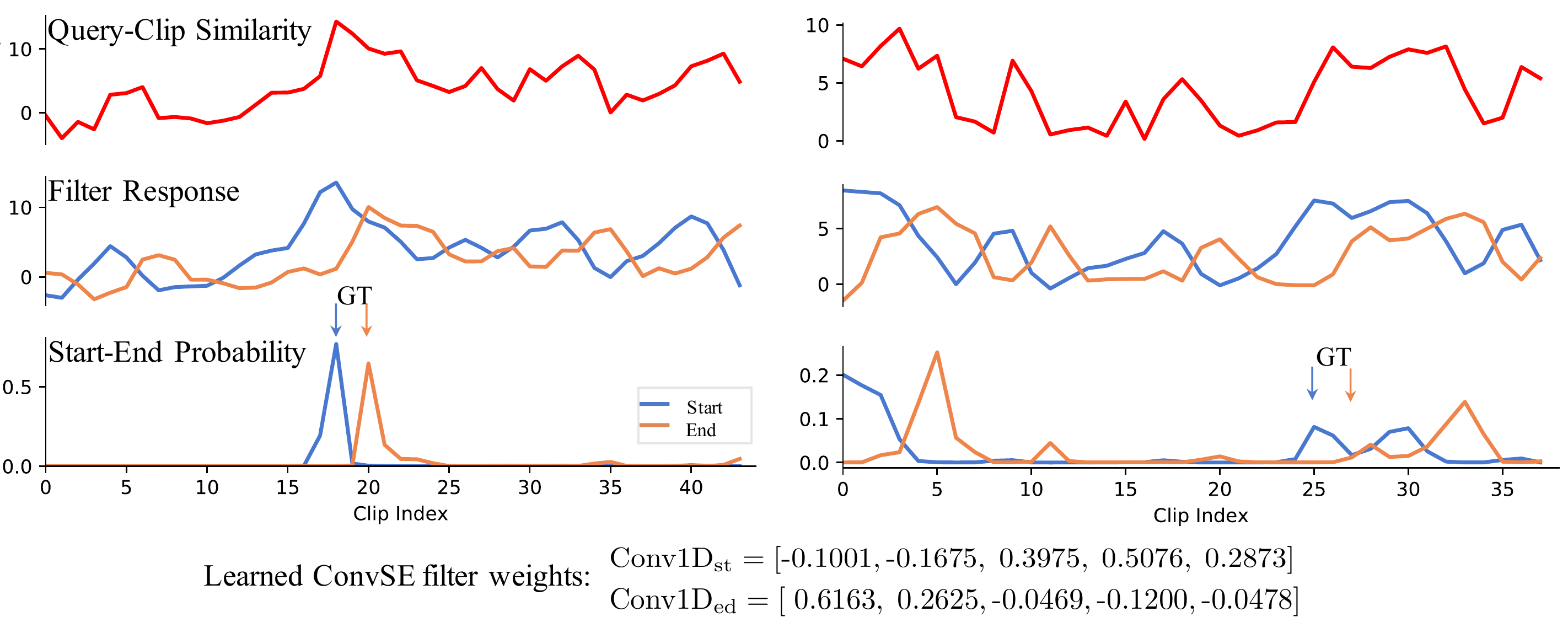}
  \caption{Examples of learned ConvSE filters applying on query-clip similarity scores. Ground truth span is indicated by the two arrows  labeled by \textit{GT}. Note the two filters output stronger responses on the up (\textit{Start}) and down (\textit{End}) edges}
  \label{fig:conv_example}
\end{figure}

In Fig.~\ref{fig:convse_comparison} (right), we vary the kernel size (\textit{k}) of ConvSE filters. While the performance is reasonable when $k\mbox{=}3$, 5 or 7, we observe a significant performance drop at $k\mbox{=}1$. 
In this case, the filters essentially degrade to scaling factors on the scores.
This comparison demonstrates that neighboring information is important.
Fig.~\ref{fig:conv_example} shows examples of using the \textbf{learned convolution filters}: the filters output stronger responses to the up (\textit{Start}) and down (\textit{End}) edges of the score curves and thus detect them.
Interestingly, the learned weights $\mathrm{Conv1D}_{\mathrm{st}}$ and $\mathrm{Conv1D}_{\mathrm{ed}}$ in Fig.~\ref{fig:conv_example} are similar to the edge detectors in image processing~\cite{szeliski2010computer}.

\begin{figure*}[!t]
  \centering
  \includegraphics[width=0.98\linewidth]{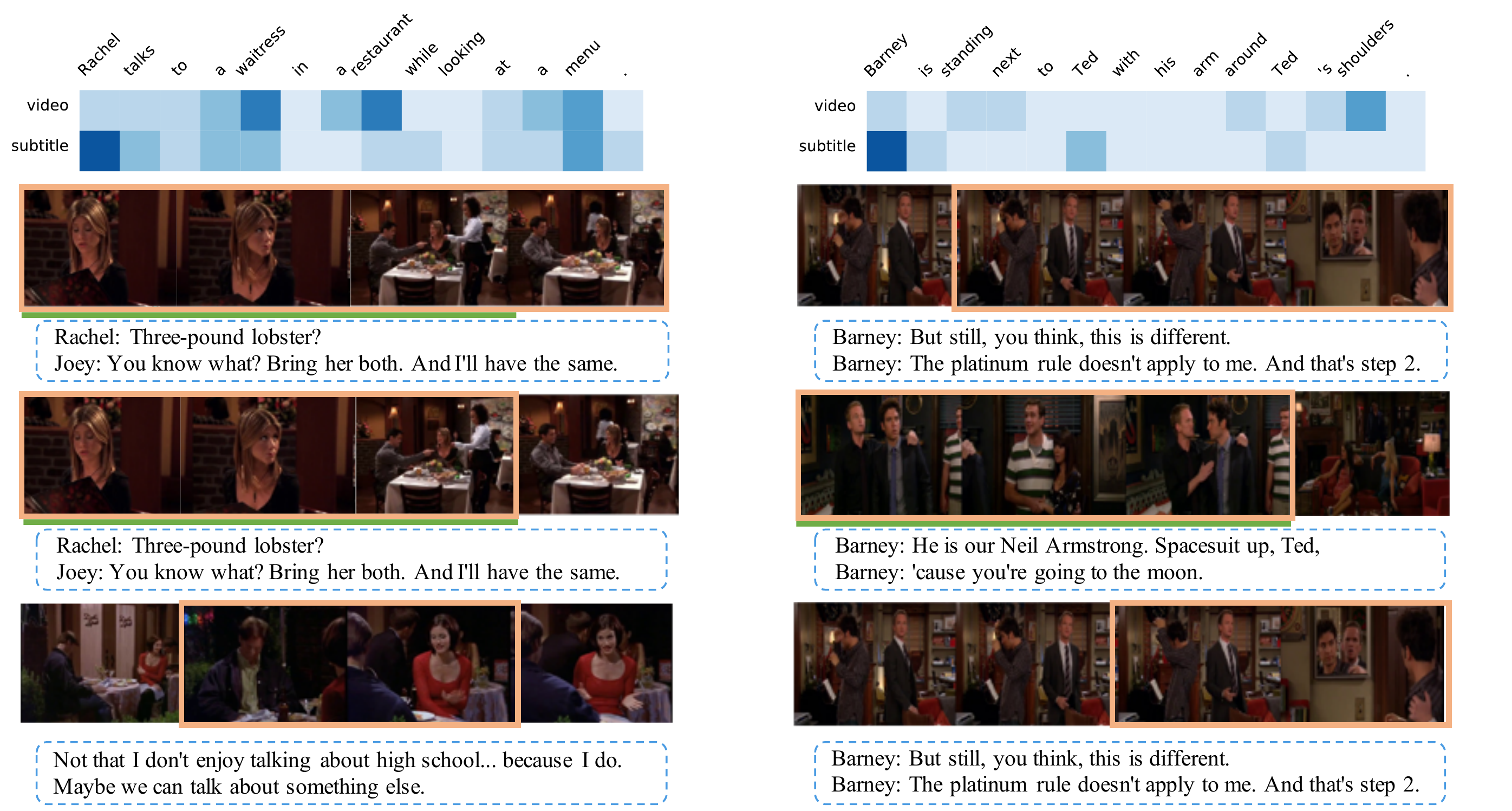}
  \caption{XML prediction examples for VCMR, on TVR \textit{val} set. We show top-3 retrieved moments for each query. \textit{Top row} shows modular attention scores for query words. \textit{Left column} shows a correct prediction, \textit{right column} shows a failure. Text inside \textit{dashed boxes} is the subtitles associated with the predicted moments. \textit{Orange box} shows the predictions, \textit{green bar} shows the ground truth}
  \label{fig:model_predictions}
\end{figure*}

\kern1mm
\noindent\textbf{Qualitative Analysis}.
Fig.~\ref{fig:model_predictions} shows XML example predictions on the TVR \textit{val} set. 
In the top row, we also show the query word attention scores for video and subtitle, respectively. Fig.~\ref{fig:model_predictions} (left) shows a correct prediction. The top-2 moments are from the same video and are both correct. The third moment is retrieved from a different video. While incorrect, it is still relevant as it also happens in a `restaurant'. Fig.~\ref{fig:model_predictions} (right) shows a failure. It is worth noting that the false moments are very close to the correct prediction with minor differences (`on the shoulder' \textit{vs.} `around the shoulder'). Besides, it is also interesting to see which words are important for video or subtitle. For example, the words `waitress', `restaurant', `menu' and `shoulder' get the most weight for video; while the words `Rachel', `menu', `Barney', `Ted' have higher attention scores for subtitle.

\section{Conclusion}\label{conclusion}
In this work, we present TVR, a large-scale dataset designed for multimodal moment retrieval tasks. Detailed analyses show TVR is of high quality and is more challenging than previous datasets. We also propose Cross-modal Moment Localization (XML), an efficient model suitable for the VCMR task.

\smallskip
\noindent
{\bf Acknowledgements:} We thank the reviewers for their helpful feedback. This research is supported by NSF Award \#1562098, DARPA MCS Grant \#N66001-19-2-4031, DARPA KAIROS Grant \#FA8750-19-2-1004, ARO-YIP Award \#W911NF-18-1-0336, and Google Focused Research Award.

\appendix
\section{Additional TVR Data Details}

\subsection{Data Collection}\label{subsec:appendix_data_collection}

\begin{figure}[!ht]
  \centering
  \includegraphics[width=\linewidth]{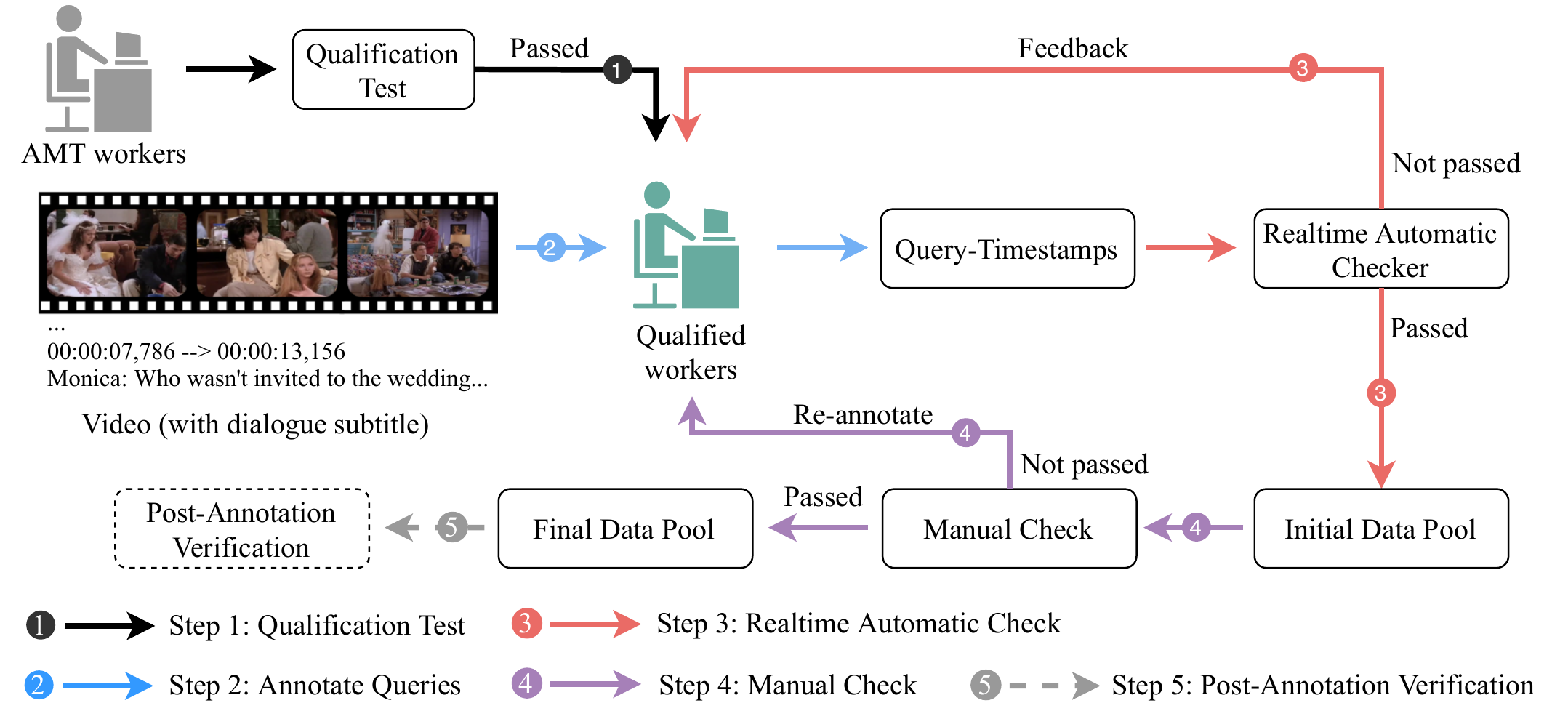}
  \caption{TVR data collection procedure}
  \label{fig:tvr_collection_overview}
\end{figure}

\noindent\textbf{TVR Data Collection Procedure.}
In Fig.~\ref{fig:tvr_collection_overview} we show an overview of TVR data collection procedure.
For details of each step, please refer to both Sec.~\ref{subsec:data_collection} and the rest of this section.

\begin{figure}[!ht]
  \centering
  \includegraphics[width=0.82\linewidth]{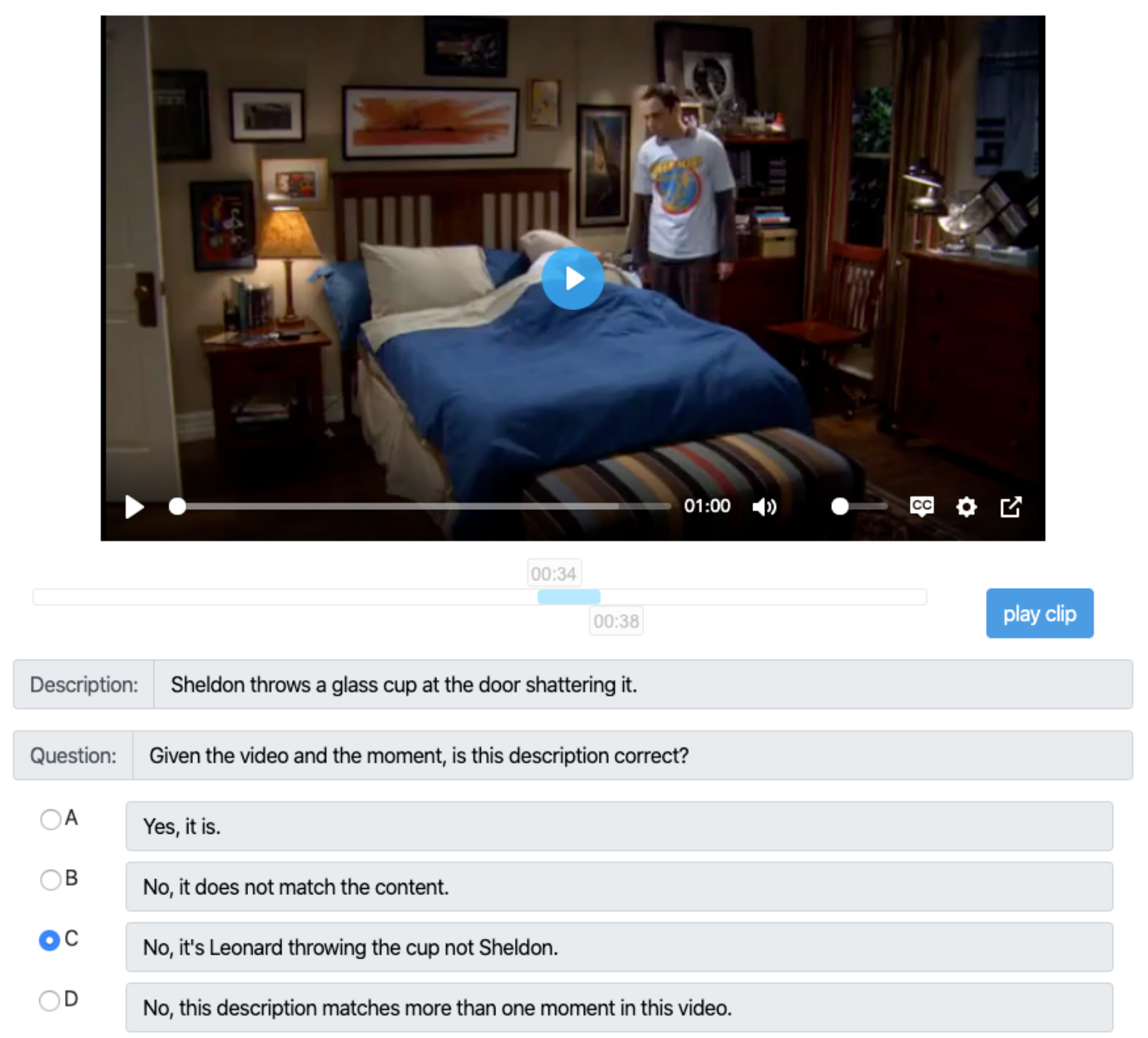}
  \caption{Example question from our qualification test}
  \label{fig:qual_test_example}
\end{figure}

\kern2mm
\noindent\textbf{Qualification Test}.
We designed a qualification test with 12 multiple-choice questions and only let workers who correctly answer at least 9 questions participate in our annotation task, ensuring that workers understand our task requirements well.
In total, 1,055 workers participated in the test, with a pass rate of 67\%. 
Adding this qualification test greatly improved data quality. In Fig.~\ref{fig:qual_test_example}, we show a question from our qualification test. 
This particular question is designed to make sure the annotators write relevant and correct descriptions (queries). 

\begin{figure}[!t]
  \centering
  \includegraphics[width=0.9\linewidth]{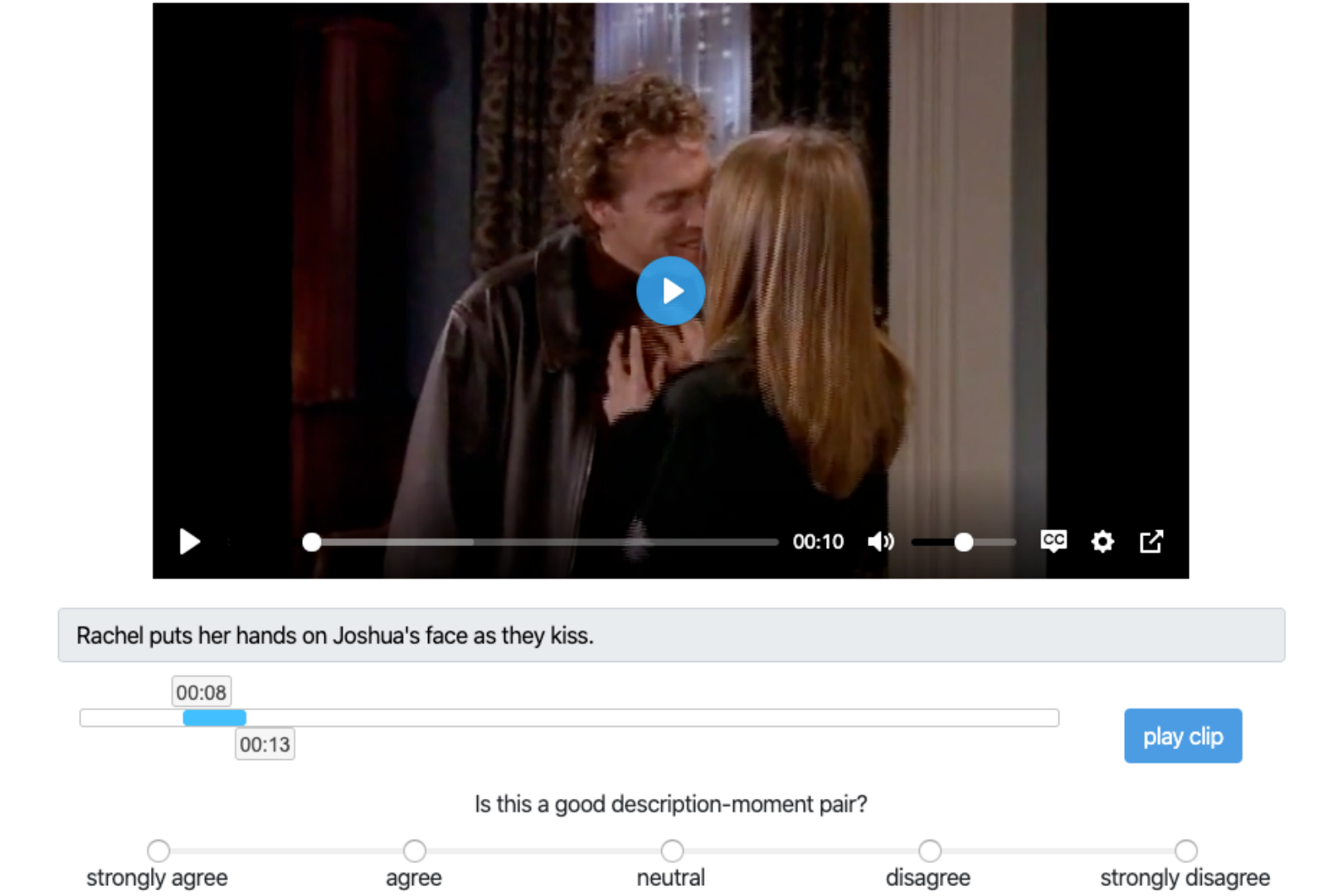}
  \caption{Post-Annotation quality rating interface}
  \label{fig:likert_scale_interface}
\end{figure}

\begin{figure}[!t]
  \centering
  \includegraphics[width=0.63\linewidth]{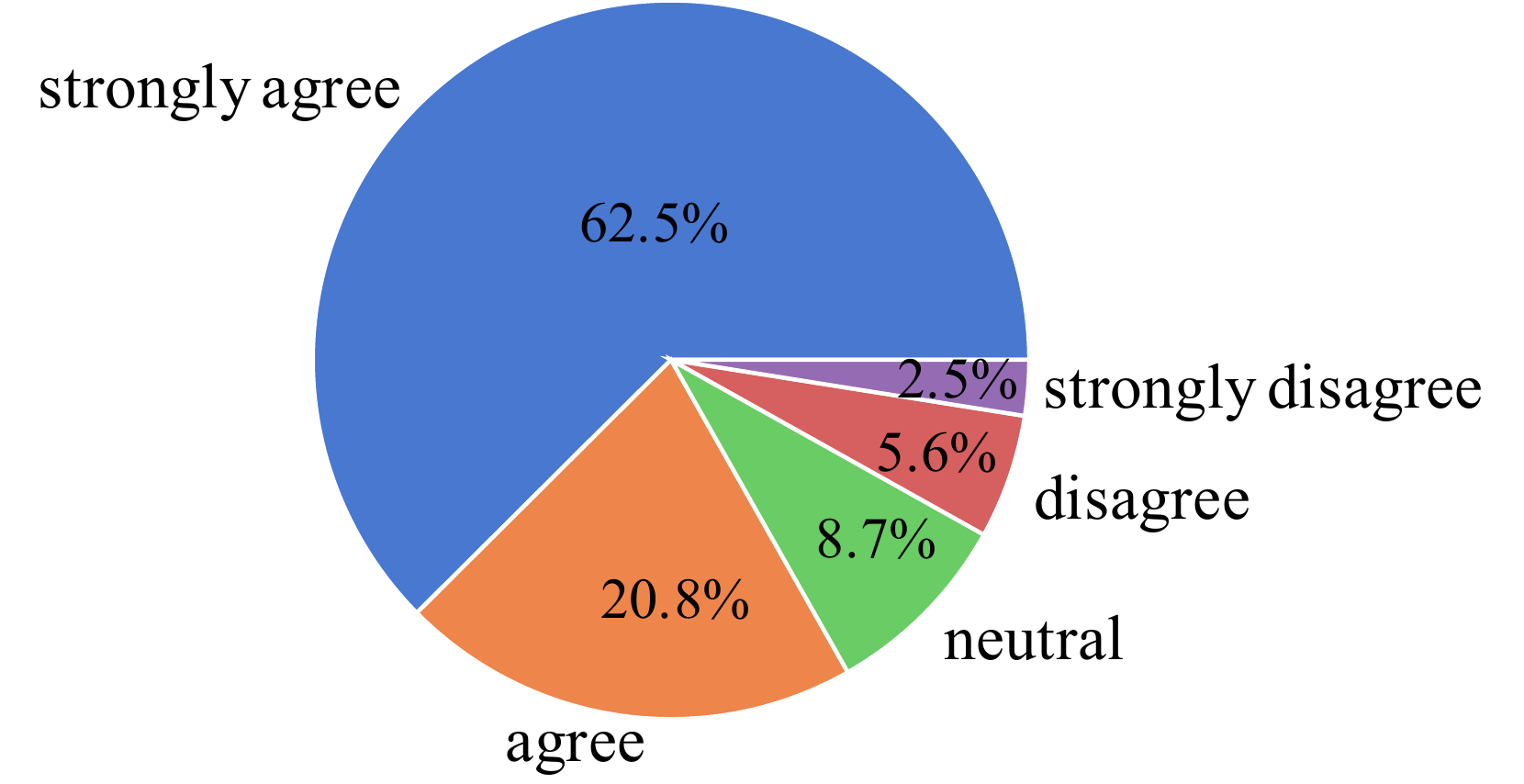}
  \caption{Distribution of quality rating on 3,600 query-moment pairs. 92\% of the pairs have a rating of at least \textit{neutral}
  }
  \label{fig:likert_rating_dist}
\end{figure}

\kern2mm
\noindent\textbf{Post-Annotation Verification}. 
To verify the quality of the collected data, we performed a post-annotation verification experiment. 
We set up another AMT task where workers were required to rate the quality of the collected query-moment pairs based on \textit{relevance}, \textit{is the query-moment pair a unique-match}, etc. 
The rating was done in a \textit{likert-scale} manner with 5 options: \textit{strongly agree}, \textit{agree}, \textit{neutral}, \textit{disagree} and \textit{strongly disagree}, as is shown in Fig.~\ref{fig:likert_scale_interface}.
Results show that 92\% of the pairs have a rating of at least \textit{neutral}. This verification was conducted on 3,600 query-moment pairs. Detailed rating distribution is shown in Fig.~\ref{fig:likert_rating_dist}. We further analyzed the group of queries that were rated as \textit{strongly disagree}, and found that 80\% of them were still of acceptable quality: e.g., slightly mismatched timestamps ($\leq$1 sec.). For the group of queries that were rated as \textit{disagree}, this number is 90\%. This verification demonstrates the high quality of the data.

\subsection{Data Analysis}\label{subsec:supp_data_analysis}
\noindent\textbf{Statistics by TV Show}. TVR is built on 21,793 videos (provided by TVQA~\cite{Lei2018TVQALC}) from 6 long-running TV shows: \textit{The Big Bang Theory}, \textit{Friends}, \textit{How I Met You Mother}, \textit{Grey's Anatomy}, \textit{House}, \textit{Castle}. Table~\ref{tab:dset_stat_by_genre} shows detailed statistics.

\begin{table}[t]
\setlength{\tabcolsep}{0.3em}
\ra{1.}
\small
\centering
\caption{Data Statistics for each TV show. BBT=\textit{The Big Bang Theory}, HIMYM=\textit{How I Met You Mother}, Grey=\textit{Grey's Anatomy}, Epi=Episode, Sea.=Season}
\scalebox{0.92}{
\begin{tabular}{l
P{0.9cm}
>{\raggedleft\arraybackslash}p{1.1cm}
>{\raggedleft\arraybackslash}p{1.2cm}
>{\raggedleft\arraybackslash}p{1.2cm}
>{\raggedleft\arraybackslash}p{1.2cm}
}
\toprule
Show    & Genre     & \#Sea. & \#Epi. & \#Clip & \#Query  \\ 
\midrule
BBT     & sitcom    & 10        & 220        & 4,198   & 20,990 \\
Friends & sitcom    & 10        & 226        & 5,337   & 26,685 \\
HIMYM   & sitcom    & 5         & 72         & 1,512   & 7,560  \\
Grey    & medical   & 3         & 58         & 1,427   & 7,135  \\
House   & medical   & 8         & 176        & 4,621   & 23,105 \\
Castle  & crime & 8         & 173        & 4,698   & 23,490 \\ 
\midrule
Total   & \quad--- & 44 & 925  &21,793 & 108,965 \\
\bottomrule
\end{tabular}
}
\label{tab:dset_stat_by_genre}
\end{table}

\begin{figure}[!t]
  \centering
  \includegraphics[width=\linewidth]{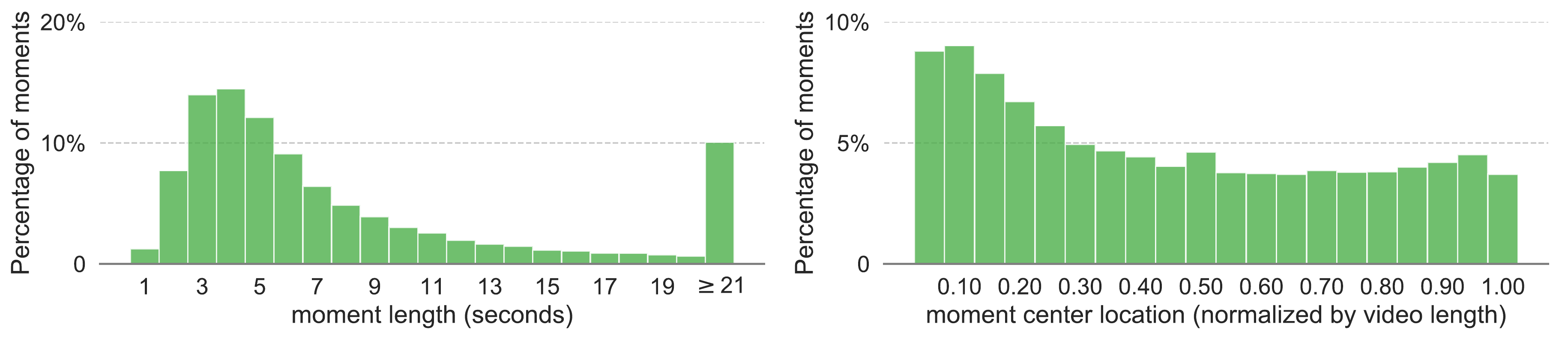}
  \caption{Distribution of TVR moment lengths (\textit{left}) and moment center locations (\textit{right})}
  \label{fig:tvr_moment_lengths_center}
\end{figure}

\begin{figure}[!t]
  \centering
  \includegraphics[width=0.78\linewidth]{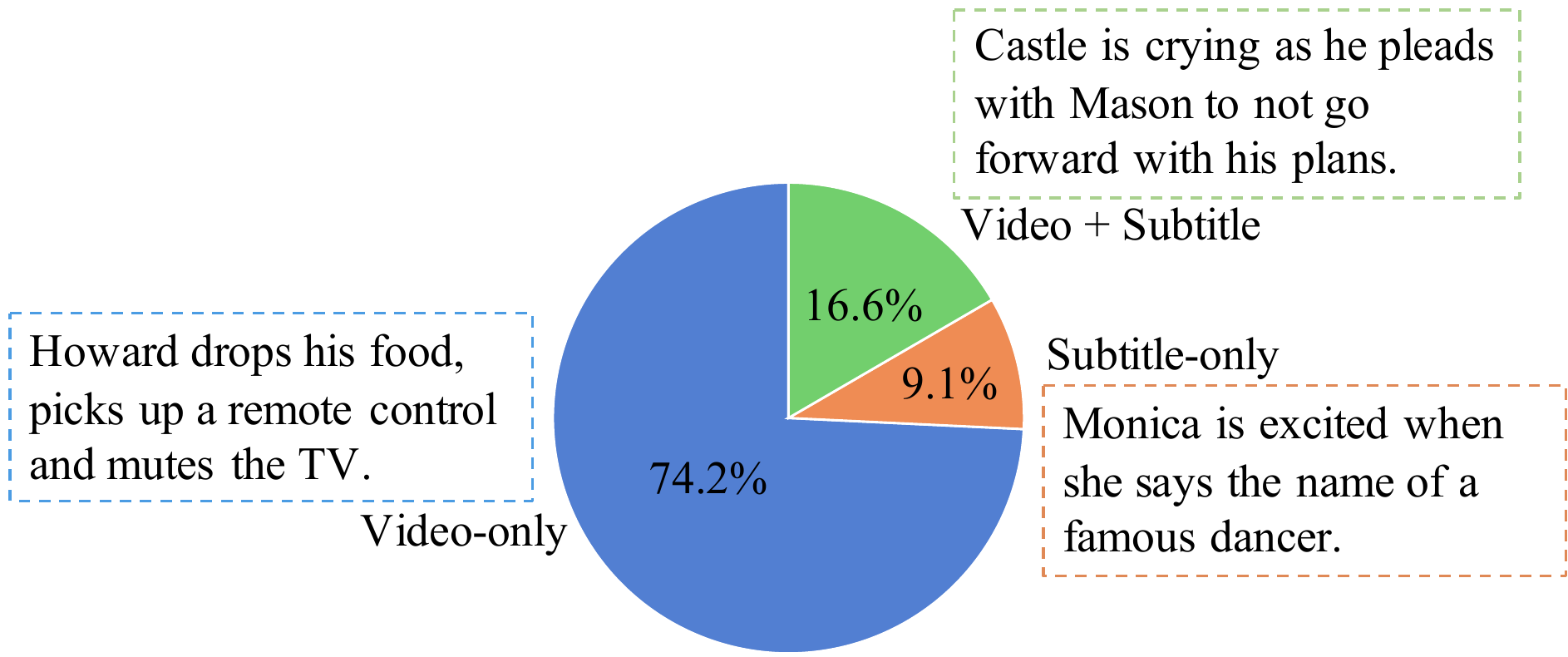}
  \caption{Distribution of query types based on reasoning type. Text inside \textit{dashed boxes} are query examples for each query type} 
  \label{fig:qtype_dist}
\end{figure}

\kern2mm
\noindent\textbf{Moments and Queries}.
Fig.~\ref{fig:tvr_moment_lengths_center} (\textit{left}) shows TVR moment length distribution. The majority of the moments are relatively short, with an average length of 9.1 secs. As a comparison, the average length of the videos is 76.2 secs. Fig.~\ref{fig:tvr_moment_lengths_center} (\textit{right}) shows the video-length normalized moment center distributions. More moments are located at the beginning of the videos. A similar phenomenon was observed in DiDeMo~\cite{anne2017localizing}. Fig.~\ref{fig:qtype_dist} shows TVR query type distribution, around 91\% of the queries need video context, while 26\% of the queries need subtitle context.

\begin{figure}[!t]
  \centering
  \includegraphics[width=0.96\linewidth]{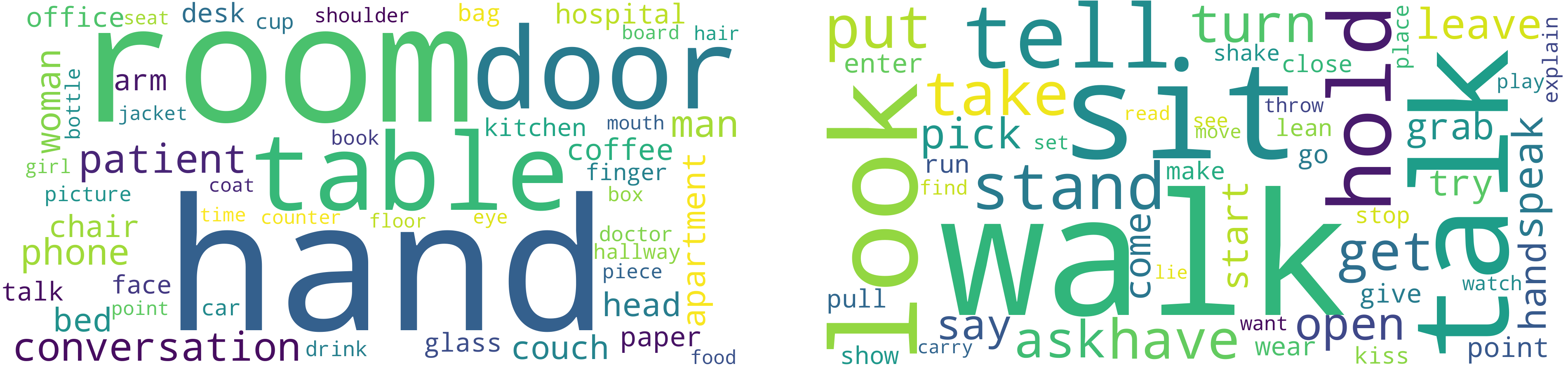}
  \caption{TVR query word clouds for nouns (\textit{left}) and verbs (\textit{right})}
  \label{fig:tvr_wordclouds}
\end{figure}

\kern2mm
\noindent\textbf{Frequent Words in Queries}.
In Fig.~\ref{fig:tvr_wordclouds} we show frequent nouns (\textit{left}) and verbs (\textit{right}) in TVR queries. The words are lemmatized, stop words are removed. We notice that TVR covers a wide range of common objects/scenes and actions, while also has many genre-specific words such as `patient' and `hospital'.

\begin{figure}[!t]
  \centering
  \includegraphics[width=0.95\linewidth]{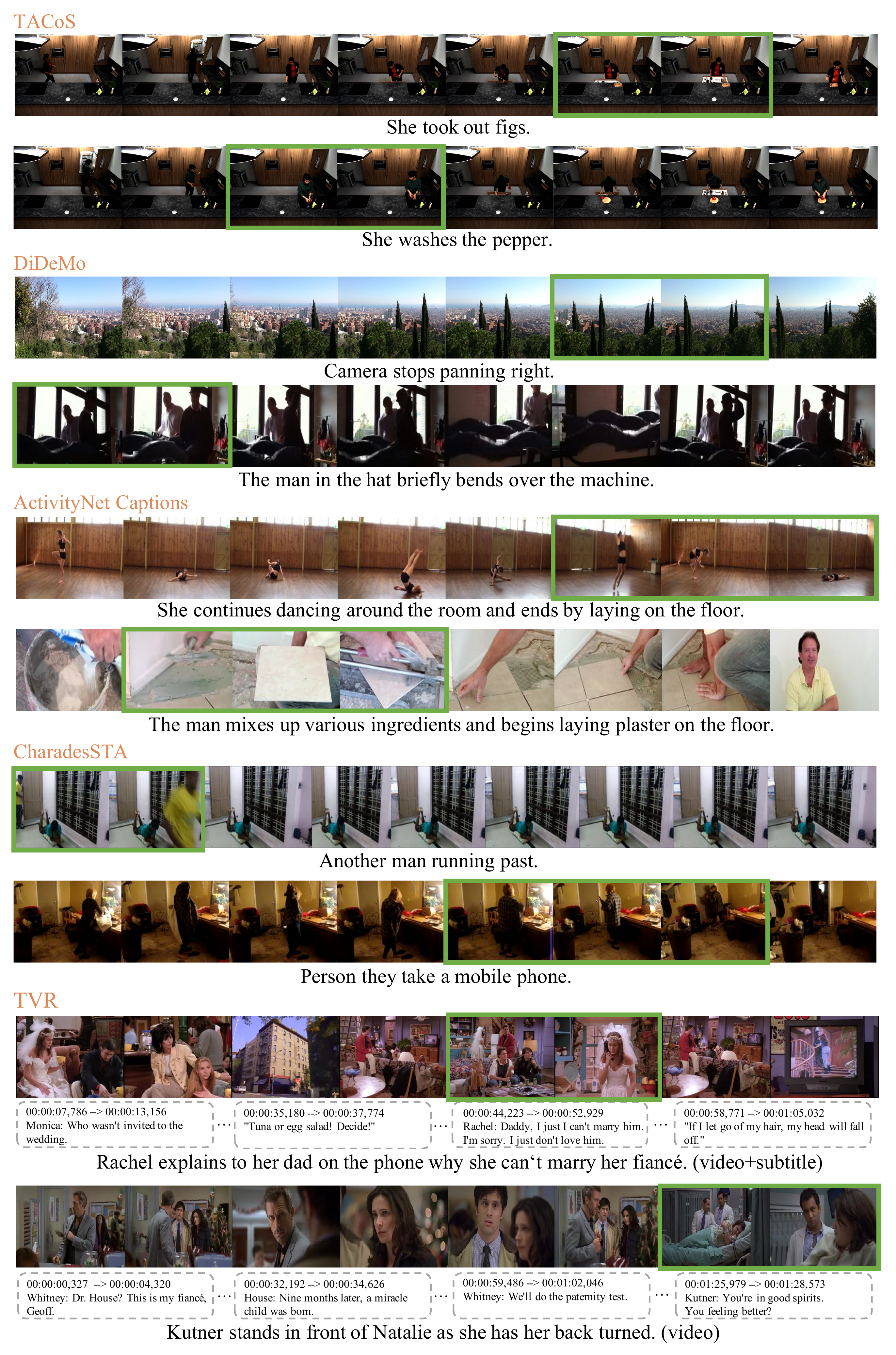}
  \caption{Comparison of TVR with existing moment retrieval datasets~\cite{regneri2013grounding,gao2017tall,Krishna2017DenseCaptioningEI,anne2017localizing}.
  Ground truth moment is shown in \textit{green box}. TVR videos are typically more diverse, containing more camera viewpoints, activities and people, etc.} 
  \label{fig:video_comparison}
\end{figure}

\kern2mm
\noindent\textbf{Video Comparison}.
TVR videos are from 6 TV shows of 3 different genres, covering a diverse set of objects/scenes/activities. In Fig.~\ref{fig:video_comparison}, we compare TVR videos with videos from existing datasets~\cite{regneri2013grounding,gao2017tall,Krishna2017DenseCaptioningEI,anne2017localizing}. 
Each TVR video typically has more visual diversity, i.e., more camera viewpoints, activities and people, etc.

\section{Additional TVR Experiments}\label{subsec:additional_tvr_experiments}

\begin{table*}[!t]
\setlength{\tabcolsep}{0.3em}
\ra{1.}
\centering
\small
\caption{Baseline comparison on TVR \textit{test-public} set, VCMR task. Model references: \textit{MCN}~\cite{anne2017localizing},
\textit{CAL}~\cite{escorcia2019temporal}, \textit{MEE}~\cite{miech2018learning}, \textit{ExCL}~\cite{ghosh2019excl}. This table includes models trained with Temporal Endpoint Feature (TEF)~\cite{anne2017localizing}}
\scalebox{0.74}{
\begin{tabular}{lP{1.1cm}P{1.2cm}
>{\raggedleft\arraybackslash}p{0.9cm}
>{\raggedleft\arraybackslash}p{0.9cm}
>{\raggedleft\arraybackslash}p{0.9cm}
>{\raggedleft\arraybackslash}p{0.9cm}
>{\raggedleft\arraybackslash}p{0.9cm}
>{\raggedleft\arraybackslash}p{0.9cm}
>{\raggedleft\arraybackslash}p{0.9cm}
>{\raggedleft\arraybackslash}p{0.9cm}
r
}
\toprule
\multirow{2}{*}{Model} & \multirow{2}{*}{w/ video} & \multirow{2}{*}{w/ sub.} & \multicolumn{4}{c}{IoU=0.5} & \multicolumn{4}{c}{IoU=0.7} & \multicolumn{1}{c}{Runtime~$\downarrow$} \\  
\cmidrule(l){4-7}  \cmidrule(l){8-11}
 & & & R@1 & R@5 & R@10 & R@100 & R@1 & R@5 & R@10 & R@100 & \multicolumn{1}{c}{(seconds)}\\ 
\midrule
Chance & - & - & 0.00 & 0.02 & 0.04 & 0.33 & 0.00 & 0.00 & 0.00 & 0.07 & - \\
Frequency & - & - & 0.06 & 0.07 & 0.11 & 0.28 & 0.02 & 0.04 & 0.06 & 0.11 & - \\ 
\multicolumn{3}{l}{\textbf{Proposal based Methods}} & & & &  &  &  & & \\
TEF-only & - & - & 0.00 & 0.09 & 0.15 & 0.79 & 0.00 & 0.07 & 0.09 & 0.48 & - \\ 
MCN & \checkmark & \checkmark & 0.02 & 0.15 & 0.24 & 2.20 & 0.00 & 0.07 & 0.09 & 1.03 & -\\
MCN (TEF) & \checkmark & \checkmark & 0.04 & 0.11 & 0.17 & 1.84 & 0.02 & 0.06 & 0.07 & 1.10 & - \\
CAL & \checkmark & \checkmark & 0.09 & 0.31 & 0.57 & 3.42 & 0.04 & 0.15 & 0.26 & 1.89 & - \\
CAL (TEF) & \checkmark & \checkmark & 0.04 & 0.17 & 0.31 & 2.48 & 0.02 & 0.15 & 0.22 & 1.30 & - \\ 
\multicolumn{3}{l}{\textbf{Retrieval + Re-ranking}} & & & &  &  &  & & \\
MEE+MCN & \checkmark & \checkmark & 0.92 & 3.69 & 5.58 & 17.91 & 0.42 & 1.89 & 2.98 & 10.84 & - \\
MEE+MCN (TEF) & \checkmark & \checkmark & 1.36 & 3.89 & 5.79 & 19.34 & 0.62 & 2.04 & 3.21 & 11.66 & 66.8 \\
MEE+CAL & \checkmark & \checkmark & 0.97 & 3.75 & 5.80 & 18.66 & 0.39 & 1.69 & 2.98 & 11.52 & - \\ 
MEE+CAL (TEF) & \checkmark & \checkmark & 1.23 & 4.00 & 6.52 & 20.07 & 0.66 & 1.93 & 3.09 & 12.03 & 161.5 \\
MEE+ExCL & \checkmark & \checkmark & 0.92 & 2.53 & 3.60 & 6.01 & 0.33 & 1.19 & 1.73 & 2.87 & - \\
MEE+ExCL (TEF) & \checkmark & \checkmark & 1.01 & 2.50 & 3.60 & 5.77 & 0.40 & 1.21 & 1.73 & 2.96 & 1307.2 \\ 
\midrule
XML (sw) & \checkmark & \checkmark & 3.82 & 10.38 & 14.20 & 35.89 & 1.91 & 5.25 & 8.12 & 23.47 & - \\
XML & \checkmark & \checkmark & \textbf{7.25} & \textbf{16.24} & \textbf{21.65} & \textbf{44.44} & \textbf{3.25} & \textbf{8.71} & \textbf{12.49} & \textbf{29.51} & - \\
XML (TEF) & \checkmark & \checkmark & \textbf{7.88} & \textbf{16.53} & \textbf{21.84} & \textbf{45.51} & \textbf{3.32} & \textbf{9.46} & \textbf{13.41} & \textbf{30.52} & \textbf{25.5} \\
\bottomrule
\end{tabular}
}
\label{tab:main_vcmr_res_tef}
\end{table*}

\subsection{More VCMR Experiments}\label{subsec:more_vcmr_cpmparison}

\noindent\textbf{Frequency Baseline}. Following prior works~\cite{anne2017localizing,escorcia2019temporal}, we first discretize the video-length normalized start-end points, then use moments with most frequent start-end points as predictions. For video retrieval, we randomly sample videos from the dataset. The results of this baseline is presented in Table~\ref{tab:main_vcmr_res_tef}. We observe this baseline has slightly better performance than chance, we hypothesize it is mainly caused by the fact that the annotators tend to annotate the first few seconds of the video~\cite{anne2017localizing}, as we have shown in Fig.~\ref{fig:tvr_moment_lengths_center} (\textit{Right}).

\kern2mm
\noindent\textbf{Models Trained with TEF}. It is shown in~\cite{anne2017localizing,escorcia2019temporal} that adding Temporal Endpoint Feature (TEF)~\cite{anne2017localizing} improves models' performance in moment retrieval tasks. In Table~\ref{tab:main_vcmr_res_tef}, we compare models trained with TEF. 
In most cases, adding TEF increases models' performance, which suggests there exists a certain degree of bias in the proposed dataset. 
This phenomenon is also observed by recent works~\cite{anne2017localizing,escorcia2019temporal} in various moment retrieval datasets, i.e., DiDeMo~\cite{anne2017localizing}, CharadesSTA~\cite{gao2017tall} and ActivityNet Captions~\cite{Krishna2017DenseCaptioningEI}. 
We attribute this phenomenon into two aspects: (1)\textit{moment distribution bias} - the moments are not evenly distributed over the video, e.g., in TVR and DiDeMo~\cite{anne2017localizing}, there are more moments appear at the beginning of the video. (2)\textit{language timestamp correlation bias} - some query words are highly indicative of the potential temporal location of the queries, e.g., temporal connectives like `first' strongly indicate the associated query might be located around the beginning of the video and pronouns like `He' may suggest this query should not be placed at the beginning of the video as people would usually not use pronouns when they first mention someone. 
The second bias commonly exists in datasets that are built by converting paragraphs into separate sentences, i.e., CharadesSTA~\cite{gao2017tall}, TACoS~\cite{regneri2013grounding} and ActivityNet Captions~\cite{Krishna2017DenseCaptioningEI}. 
TVR avoids this bias by explicitly ask annotators to write queries as individual sentences without requiring the context of a paragraph.

\kern2mm
\noindent\textbf{XML with Sliding Windows}.
In Sec.~\ref{subsec:model_analysis}, we compared XML variants with different proposal generation strategies. 
In Table~\ref{tab:main_vcmr_res_tef}, we further compare XML (sw, sliding window) with MCN/CAL models. For details of this variant, see Sec.~\ref{subsec:model_analysis}.
Compared to the best baseline (MEE+CAL), using the same set of sliding window proposals, we observe XML (sw) still perform much better (3.82 \textit{vs.} 0.97, R@1 IoU=0.7).
We hypothesize that the lower performance of MCN/CAL models compared to XML is mainly caused by the difficulties of training and ranking with a large pool of proposal candidates (1.5M proposals for TVR \textit{train}). Both MCN and CAL are trained with a ranking objective, which relies on informative negatives to learn effectively. However, effective negative sampling in such a large pool of candidates can be challenging. In comparison, XML breaks the video corpus level moment retrieval problem into two sub-problems: video-level and moment-level retrieval. At video-level retrieval, XML performs ranking within a small set of videos (17.4K), which eases the aforementioned issue. At moment-level, XML (sliding window) utilizes Binary Cross Entropy to maximize the similarity scores of each ground-truth clip, eliminating the need for manually designing a negative sampling strategy.

\begin{table}[!t]
\setlength{\tabcolsep}{0.3em}
\ra{1.}
\centering
\small
\caption{Model architecture ablation on TVR \textit{val} set, VCMR task. Our full XML model in the last row is configured with transformer encoder and modular query. All models use both videos and subtitles}
\scalebox{0.86}{
\begin{tabular}{p{4cm}
>{\raggedleft\arraybackslash}p{0.9cm}
>{\raggedleft\arraybackslash}p{0.9cm}
>{\raggedleft\arraybackslash}p{0.9cm}
>{\raggedleft\arraybackslash}p{0.9cm}
}
\toprule
\multirow{2}{*}{Model} & \multicolumn{4}{c}{IoU=0.7} \\  
\cmidrule(l){2-5}
& R@1 & R@5 & R@10 & R@100\\ 
\midrule
\multicolumn{3}{l}{\textbf{Self-Encoder Type}}  & &   \\
XML (LSTM)   & 2.12 & 4.97 & 6.86 & 18.06 \\
XML (CNN) & 2.45 & 5.53 & 7.77 & 19.88 \\
\multicolumn{3}{l}{\textbf{Modular Query}} & & \\
XML (No modular query) & 2.46 & 5.87 & 8.56 & 22.00 \\ 
\midrule
XML  & \textbf{2.62} & \textbf{6.39} & \textbf{9.05} & \textbf{22.47} \\
\bottomrule
\end{tabular}
}
\label{tab:ablation_model}
\end{table}

\kern1mm
\noindent\textbf{Model Architecture}.
Table~\ref{tab:ablation_model} presents a model architecture ablation. 
We first compare with different self-encoder architectures, replacing our transformer style encoder with a bidirectional LSTM encoder~\cite{Lei2018TVQALC} or a CNN encoder~\cite{yu2018qanet,lei2019tvqa+}. 
We observe worse performance after the change and attribute this performance drop to the ineffectiveness of LSTMs and CNNs to capture long-term dependencies~\cite{hochreiter2001gradient,vaswani2017attention}. 
Next, we compare XML with a variant that uses a single max-pooled query instead of two modularized queries. Across all metrics, XML performs better than the variant without modular queries, showing the importance of considering different query representations in matching the context from different modalities.

\begin{table}[!t]
\setlength{\tabcolsep}{0.3em}
\ra{1.}
\centering
\small
\caption{Feature ablation on TVR \textit{val} set, VCMR task. All  models use both videos and subtitles}
\scalebox{0.86}{
\begin{tabular}{p{3cm}
>{\raggedleft\arraybackslash}p{0.9cm}
>{\raggedleft\arraybackslash}p{0.9cm}
>{\raggedleft\arraybackslash}p{0.9cm}
>{\raggedleft\arraybackslash}p{0.9cm}
}
\toprule
\multirow{2}{*}{Model}  & \multicolumn{4}{c}{IoU=0.7} \\  
\cmidrule(l){2-5} 
&  R@1 & R@5 & R@10 & R@100 \\ 
\midrule
XML (ResNet)  & 2.28 & 5.40 & 7.33 & 20.28 \\
XML (I3D) & 2.22 & 5.75 & 8.37 & 21.20 \\
XML (ResNet+I3D)  & \textbf{2.62} & \textbf{6.39} & \textbf{9.05} & \textbf{22.47} \\
\bottomrule
\end{tabular}
}
\label{tab:ablation_feature}
\end{table}

\kern2mm
\noindent\textbf{Feature Ablation}. We tested XML model with different visual features, the results are shown in Table~\ref{tab:ablation_feature}. 
The model that uses both static appearance features (ResNet~\cite{he2016deep}) and action features (I3D~\cite{carreira2017quo}) outperforms models using only one of the features, demonstrating the importance of recognizing both the objects and the actions in the VCMR task.

\kern2mm
\noindent\textbf{Retrieval Efficiency in 1M Videos}. We consider Video Corpus Moment Retrieval in a video corpus containing 1M videos with 100 queries. Following~\cite{escorcia2019temporal}, we conduct this experiment in a simulated setting with each video containing 20 clips with max moment length of 14 clips. Each query containing 15 words. We report the following metrics: (1) feature encoding time (\textit{feat time}) - measures the time for encoding the context (video and subtitle) features offline. (2) encoded feature size (\textit{feat size}) - measures the disk space needed to store the encoded context features. (3) retrieval time (\textit{retrieval time}) - measures the time needed to retrieve relevant moments for 100 new queries. It includes time for encoding the queries and performing approximate nearest neighbor search~\cite{JDH17} or matrix multiplication. The time spent on data loading, pre-processing, feature extraction on backend models (i.e., ResNet-152, I3D, RoBERTa) are not considered as they should be similar if not the same for all the methods. Note that the \textit{retrieval time} here is different from the \textit{runtime} in Table~\ref{tab:main_vcmr_res_tef}, which additional includes \textit{feat time}. We do not report \textit{feat time} and \textit{feat size} for ExCL~\cite{ghosh2019excl} as it does not have the ability to pre-encode the features - its context encoding depends on the input queries. This experiment was conducted on an RTX 2080Ti GPU and an Intel(R) Xeon(R) Silver 4114 CPU @ 2.20GHz $\times$ 40, with PyTorch~\cite{paszke2017automatic} and FAISS~\cite{JDH17}.

\begin{table}[!t]
\setlength{\tabcolsep}{0.3em}
\ra{1.}
\centering
\small
\caption{VCMR on 1M videos with 100 queries. TVR \textit{test-public} set results are included as reference.Model references: \textit{MCN}~\cite{anne2017localizing},
\textit{CAL}~\cite{escorcia2019temporal}, \textit{MEE}~\cite{miech2018learning}, \textit{ExCL}~\cite{ghosh2019excl}}

\scalebox{0.82}{
\begin{tabular}{
p{4cm}
p{0.9cm}
p{0.9cm}
>{\raggedleft\arraybackslash}p{2cm}
>{\raggedleft\arraybackslash}p{2cm}
>{\raggedleft\arraybackslash}p{2.5cm}
}
\toprule
\multirow{2}{*}{Model} & \multicolumn{2}{c}{IoU=0.7} & \multicolumn{3}{c}{Search 100 queries in 1M videos $\downarrow$} \\  
\cmidrule(r){2-3} \cmidrule(l){4-6} 
& R@1 & R@5 & feat time (s) & feat size (GB)  & retrieval time (s) \\ 
\midrule
\multicolumn{3}{l}{\textbf{Retrieval + Re-ranking}} &  & \\
MEE+MCN & 0.42 & 1.89 & 131  & 326 & 0.090\\
MEE+CAL &  0.39 & 1.69  & 841 & 2,235 & 0.166 \\
MEE+ExCL &  0.33 & 1.19 & - & -  & 1.435 \\ 
\midrule
XML & \textbf{3.25} & \textbf{8.71} & \textbf{29}  & \textbf{76} & \textbf{0.005}\\
\bottomrule
\end{tabular}
}
\label{tab:retrieval_1m}
\end{table}

The results are shown in Table~\ref{tab:retrieval_1m}. Our XML model is more efficient than all the baselines. Compared to the best baseline methods MEE+MCN, XML is $18\times$ faster in retrieval, $4.5\times$ faster in feature encoding and needs 77\% less disk space to store the encoded features. Besides, it also has $7.7\times$ higher performance (3.25 \textit{vs.} 0.42, IoU=0.7, R@1, on TVR \textit{test-public} set). Note that MEE+ExCL has very poor \textit{retrieval time} performance ($287\times$ slower than XML), as it requires early fusion of context and query features. In comparison, the other 3 methods are able to pre-encode the context features and only perform lightweight query encoding and highly optimized nearest neighbor search or matrix multiplication to obtain the moment predictions.

\begin{table*}[!t]
\setlength{\tabcolsep}{0.3em}
\ra{1.}
\centering
\small
\caption{Impact of \#retrieved videos on TVR \textit{val} set, VCMR task.}
\scalebox{0.8}{
\begin{tabular}{lP{1.8cm}
>{\raggedleft\arraybackslash}p{0.9cm}
>{\raggedleft\arraybackslash}p{0.9cm}
>{\raggedleft\arraybackslash}p{0.9cm}
>{\raggedleft\arraybackslash}p{0.9cm}
>{\raggedleft\arraybackslash}p{0.9cm}
>{\raggedleft\arraybackslash}p{0.9cm}
>{\raggedleft\arraybackslash}p{0.9cm}
>{\raggedleft\arraybackslash}p{0.9cm}
}
\toprule
\multirow{2}{*}{Model} & \multirow{2}{*}{\#retrieved}  & \multicolumn{4}{c}{IoU=0.5} & \multicolumn{4}{c}{IoU=0.7} \\  
\cmidrule(l){3-6}  \cmidrule(l){7-10}
 & videos &  R@1 & R@5 & R@10 & R@100 & R@1 & R@5 & R@10 & R@100 \\ 
\midrule
\multirow{4}{*}{XML} & 10 & 5.29 & 11.82 & 15.83 & 31.05 & 2.62 & 6.54 & 9.14 & 21.19 \\
 & 50 & 5.29 & 11.74 & 15.92 & 35.95 & 2.63 & 6.40 & 9.07 & 22.55 \\
 & 100 & 5.28 & 11.73 & 15.90 & 36.16 & 2.62 & 6.39 & 9.05 & 22.47 \\
 & 200 & 5.28 & 11.73 & 15.90 & 36.20 & 2.62 & 6.39 & 9.05 & 22.46 \\
\bottomrule
\end{tabular}
}
\label{tab:vcmr_num_video}
\end{table*}

\kern2mm
\noindent\textbf{Impact of \#Retrieved Videos}. In previous experiments, we fix the number of videos retrieved by XML to be 100 for corpus level moment retrieval experiments. To study the impact of this hyperparameter, we perform experiments when \#videos $\in [10, 50, 100, 200]$, the results are shown in Table~\ref{tab:vcmr_num_video}. Overall, we notice XML is not sensitive to the number of retrieved videos in terms of R@1, R@5 and R@10 (IoU=0.5, 0.7) in the tested range. When we focus on R@100, IoU=0.5, we find that using more videos helps improve the retrieval performance.

\subsection{SVMR and Video Retrieval Experiments}

\begin{table}[!t]
\setlength{\tabcolsep}{0.3em}
\ra{1.}
\centering
\small
\caption{SVMR results on TVR \textit{val} set. Model references: \textit{MCN}~\cite{anne2017localizing},
\textit{CAL}~\cite{escorcia2019temporal}, \textit{MEE}~\cite{miech2018learning}, \textit{ExCL}~\cite{ghosh2019excl}. We show top-2 scores in each column in bold
}
\scalebox{0.8}{
\begin{tabular}{lP{1.4cm}P{1.4cm}%
>{\raggedleft\arraybackslash}p{0.9cm}%
>{\raggedleft\arraybackslash}p{0.9cm}%
>{\raggedleft\arraybackslash}p{0.9cm}%
>{\raggedleft\arraybackslash}p{0.9cm}%
}
\toprule
\multirow{2}{*}{Model} & \multirow{2}{*}{w/ video} & \multirow{2}{*}{w/ sub.} & \multicolumn{2}{c}{IoU=0.5} & \multicolumn{2}{c}{IoU=0.7} \\  
\cmidrule(rl){4-5}  \cmidrule(rl){6-7}
& & & R@1 & R@5 & R@1 & R@5 \\ 
\midrule
Chance & - & - & 3.24 & 12.79 & 0.94 & 4.41 \\
Moment Frequency & - & - & 7.72 & 18.93 & 4.19 & 12.27 \\ 
\midrule
TEF-only & - & - & 9.63 & 24.86 & 5.14 & 14.92 \\
MCN & \checkmark & \checkmark & 13.08 & 39.61 & 5.06 & 20.37 \\
MCN (TEF) & \checkmark & \checkmark & 16.86 & 40.55 & 7.96 & 21.45 \\
CAL & \checkmark & \checkmark & 12.07 & 39.52 & 4.68 & 20.17 \\
CAL (TEF) & \checkmark & \checkmark & 17.61 & 42.08 & 8.07 & 21.40 \\ 
ExCL & \checkmark & \checkmark & \textbf{31.34} & 47.40 & \textbf{14.19} & 28.01 \\
ExCL (TEF) & \checkmark & \checkmark & 31.31 & 48.54 & \textbf{14.34} & 28.89 \\ 
\midrule
XML & \checkmark & \checkmark & 30.75 & \textbf{51.20} & 13.41 & \textbf{31.11} \\
XML (TEF) & \checkmark & \checkmark & \textbf{31.43} & \textbf{51.66} & 13.89 & \textbf{31.11} \\
\bottomrule
\end{tabular}
}
\label{tab:main_svmr_res}
\end{table}

\begin{table}[!t]
\setlength{\tabcolsep}{0.3em}
\ra{1.}
\centering
\small
\caption{Video retrieval results on TVR \textit{val} set. Model references: \textit{MCN}~\cite{anne2017localizing},
\textit{CAL}~\cite{escorcia2019temporal}, \textit{MEE}~\cite{miech2018learning}
}
\scalebox{0.86}{
\begin{tabular}{lP{1.4cm}P{1.4cm}
>{\raggedleft\arraybackslash}p{0.9cm}
>{\raggedleft\arraybackslash}p{0.9cm}
>{\raggedleft\arraybackslash}p{0.9cm}
>{\raggedleft\arraybackslash}p{0.9cm}
}
\toprule
Model & w/ video & w/ sub. & R@1 & R@5 & R@10 & R@100 \\ 
\midrule
Chance & - & - & 0.03 & 0.22 & 0.47 & 4.61 \\ 
\midrule
MCN & \checkmark & \checkmark & 0.05 & 0.38 & 0.66 & 3.59 \\
MCN (TEF) & \checkmark & \checkmark & 0.07 & 0.28 & 0.51 & 3.93 \\
CAL & \checkmark & \checkmark & 0.28 & 1.02 & 1.68 & 8.55 \\
CAL (TEF) & \checkmark & \checkmark & 0.06 & 0.34 & 0.63 & 5.26 \\
MEE & \checkmark & \checkmark & 7.56 & 20.78 & 29.88 & 73.07 \\ 
\midrule
XML & \checkmark & \checkmark & \textbf{16.54} & \textbf{38.11} & \textbf{50.41} & \textbf{88.22} \\
XML (TEF) & \checkmark & \checkmark & \textbf{16.08} & \textbf{37.92} & \textbf{50.38} & \textbf{88.62} \\
\bottomrule
\end{tabular}
}
\label{tab:main_vr_res}
\end{table}

\kern2mm
\noindent\textbf{Single Video Moment Retrieval}.
Table~\ref{tab:main_svmr_res} shows the Single Video Moment Retrieval (SVMR) results on TVR \textit{val} set. 
The goal of the task is to retrieve relevant moments from a single video rather than from a video corpus as in VCMR.
We observe XML achieves comparable performance with the state-of-the-art method ExCL~\cite{ghosh2019excl}. 
However, note that XML significantly outperforms ExCL on the VCMR task with higher efficiency, as stated in Sec.~\ref{subsec:baseline_comparison} and Sec.~\ref{subsec:more_vcmr_cpmparison}. We also noticed that adding TEF has minimal impact on the performance of XML and ExCL, while greatly improves MCN's and CAL's performance. This is not surprising as XML and ExCL directly model the complete video where the temporal information could be acquired, while MCN and CAL break the video into separate proposals where the temporal information is lost in the process.

\kern2mm
\noindent\textbf{Video Retrieval}.
Table~\ref{tab:main_vr_res} shows the Video Retrieval results on TVR \textit{val} set. The goal of the task is to retrieve relevant videos from a large corpus.
As MCN and CAL do not perform whole-video retrieval, we approximate their video retrieval predictions using the videos associated with the top-retrieved moments, as in~\cite{escorcia2019temporal}.
MCN and CAL models perform rather poor ($>$50x lower performance than XML, R@1) on the video retrieval task, we summarize some possible reasons here: (1) MCN and CAL's video retrieval results are only an approximation as they are trained to differentiate moments rather than videos; (2) they need to rank a large number of proposals (187K proposals in TVR \textit{val} set), which has many drawbacks, e.g., inefficient negative sampling in training.
MEE gets less than half of XML's performance as it uses global pooled context features instead of more fine-grained local context features as XML.

\subsection{More Qualitative Examples}
\begin{figure*}[!t]
  \centering
  \includegraphics[width=\linewidth]{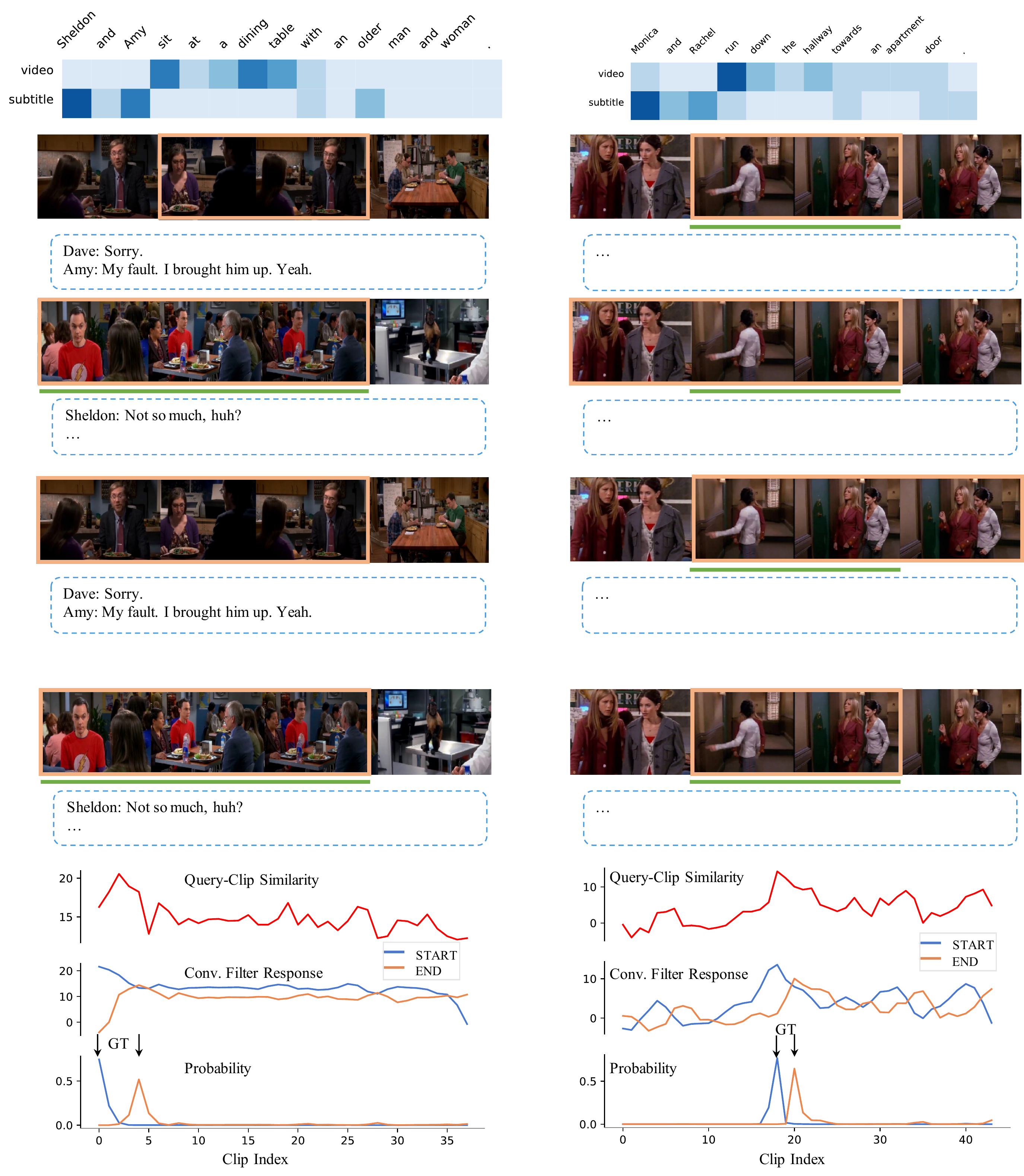}
  \caption{Qualitative examples of XML. We show top-3 retrieved moments for VCMR (\textit{top}) and SVMR results (\textit{bottom}, with convolution filter responses) for each query. Text inside \textit{dashed boxes} is the subtitles with the predicted moments. \textit{Orange box} shows the predictions, \textit{green bar} shows the ground truth. Best viewed in color}
  \label{fig:tvr_model_predictions1}
\end{figure*}

\begin{figure*}[!t]
  \centering
  \includegraphics[width=\linewidth]{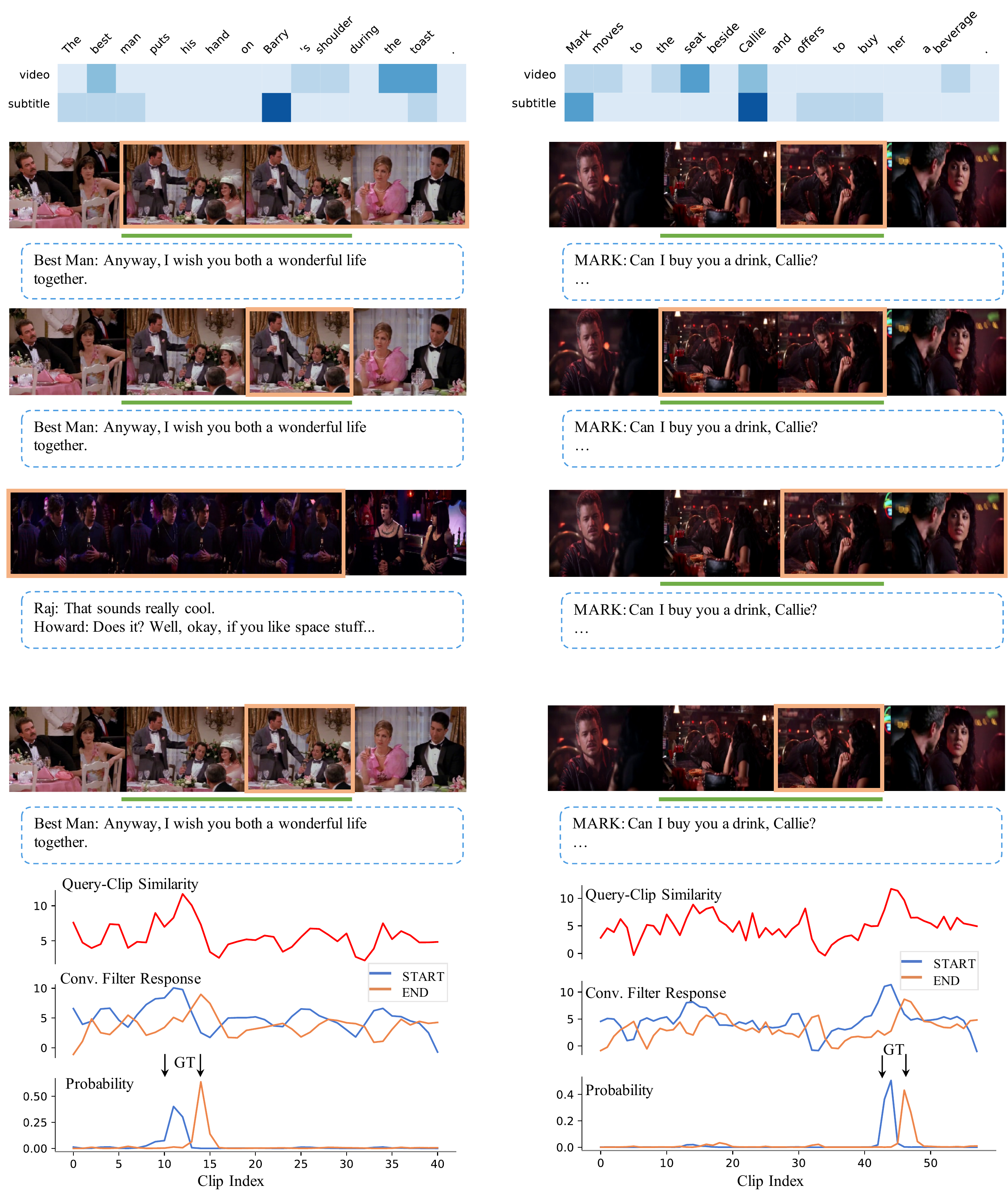}
  \caption{Qualitative examples of XML. We show top-3 retrieved moments for VCMR (\textit{top}) and SVMR results (\textit{bottom}, with convolution filter responses) for each query. Text inside \textit{dashed boxes} is the subtitles with the predicted moments. \textit{Orange box} shows the predictions, \textit{green bar} shows the ground truth. Best viewed in color}
  \label{fig:tvr_model_predictions2}
\end{figure*}

We show more qualitative examples from our XML model in Fig.~\ref{fig:tvr_model_predictions1} and Fig.~\ref{fig:tvr_model_predictions2}. We show top-3 predictions for the VCMR task, as well as associated predictions (with ConvSE filter responses) for the SVMR task.

\begin{table*}[!t]
\setlength{\tabcolsep}{0.3em}
\ra{1.}
\centering
\small
\caption{VCMR results on DiDeMo~\cite{anne2017localizing} \textit{test} set. Model references: \textit{MCN}~\cite{anne2017localizing},
\textit{CAL}~\cite{escorcia2019temporal}, \textit{MEE}~\cite{miech2018learning}. This table includes models trained with Temporal Endpoint Feature (TEF)~\cite{anne2017localizing}. We show top scores in each column in bold}
\scalebox{0.86}{
\begin{tabular}{lP{1.4cm}
>{\raggedleft\arraybackslash}p{0.9cm}
>{\raggedleft\arraybackslash}p{0.9cm}
>{\raggedleft\arraybackslash}p{0.9cm}
>{\raggedleft\arraybackslash}p{0.9cm}
>{\raggedleft\arraybackslash}p{0.9cm}
>{\raggedleft\arraybackslash}p{0.9cm}
}
\toprule
\multirow{2}{*}{Model} & \multirow{2}{*}{w/ video}  & \multicolumn{3}{c}{IoU=0.5} & \multicolumn{3}{c}{IoU=0.7}  \\  
\cmidrule(l){3-5}  \cmidrule(l){6-8}
 & & R@1 & R@10 & R@100 & R@1 & R@10 & R@100 \\ 
\midrule
Chance & - & 0.00 & 0.10 & 1.99 & 0.00 & 0.02 & 0.64 \\
Frequency & - & 0.02 & 0.22 & 2.34 & 0.02 & 0.17 & 1.99 \\
\multicolumn{3}{l}{\textbf{Proposal based Methods}} & & & &  &  \\
TEF-only & -  & 0.05 & 0.32 & 2.58 & 0.03 & 0.27 & 2.12 \\
MCN (TEF) & \checkmark & 0.88 & 5.16 & 26.23 & 0.58 & 4.12 & 21.03 \\
CAL (TEF) & \checkmark  & 0.97 & 6.15 & 28.06 & 0.66 & 4.69 & 22.89 \\
\multicolumn{3}{l}{\textbf{Retrieval + Re-ranking}} & & & &  &  \\
MEE+MCN (TEF) & \checkmark &  0.53 & 3.00 & 6.52 & 0.46 & 2.64 & 6.37 \\
MCN+MCN (TEF) & \checkmark & 0.92 & 4.83 & 17.50 & 0.64 & 3.67 & 13.12 \\
CAL+CAL (TEF) & \checkmark  & 1.07 & 6.45 & 22.60 & 0.72 & 4.86 & 17.60 \\
CAL+CAL (TEF,re-train) & \checkmark & 1.29 & 6.71 & 22.51 & 0.85 & 4.95 & 17.73 \\
Approx. CAL+CAL (TEF,re-train) & \checkmark & 1.27 & 6.39 & 15.82 & 0.80 & 4.95 & 11.59 \\
\midrule
XML (TEF) & \checkmark &  \bf 2.26 & \bf 10.42 & \bf 34.49 & \bf 1.59 & \bf 6.71 & \bf 25.44 \\
\bottomrule
\end{tabular}
}
\label{tab:main_vcmr_res_didemo}
\end{table*}

\section{TVR DiDeMo Experiments}\label{didemo_experiments}
To show the effectiveness of XML for the VCMR task, we also tested it on the popular moment retrieval dataset DiDeMo~\cite{anne2017localizing}. 
Different from TVR experiments, we only use ResNet features for DiDeMo. 
Besides, we also switch off the subtitle stream as DiDeMo has only video context.
The results are shown in Table~\ref{tab:main_vcmr_res_didemo}. 
The baseline results are directly taken from~\cite{escorcia2019temporal}. 
We observe XML outperforms all the baseline methods on DiDeMo dataset by a large margin, showing XML is able to generalize well to datasets where only video is available.

\section{TVC Dataset and Experiments}\label{sec:tvc}

\begin{figure}[!t]
  \centering
  \includegraphics[width=0.72\linewidth]{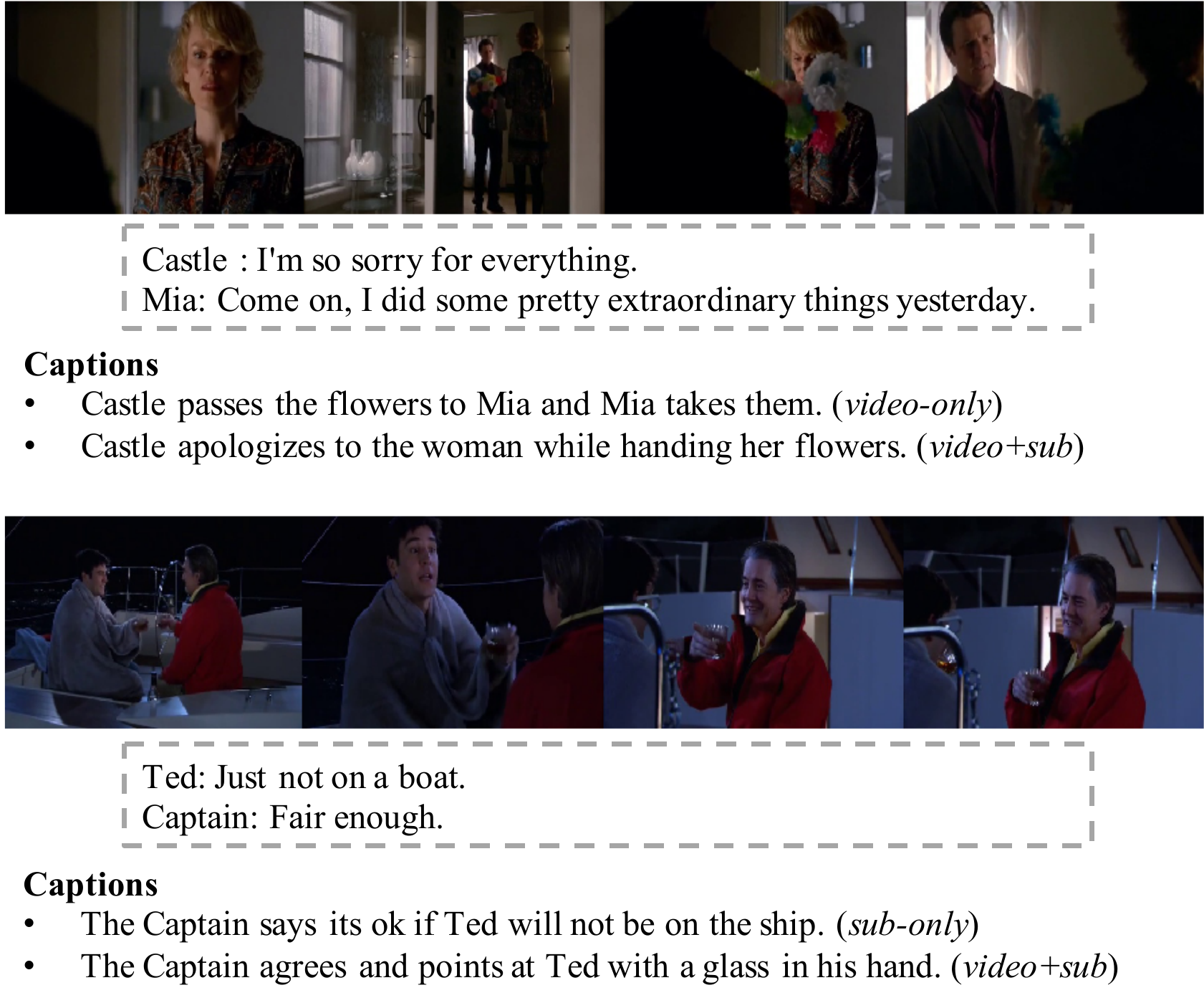}
  \caption{TVC caption description examples. Each caption description is followed by a \textit{description type tag}. Text inside \textit{dashed boxes} is the subtitles associated with the moments. For brevity, here we only show sampled frames from the moments}
  \label{fig:tvc_examples}
\end{figure}

After the TVR data collection, we extended TVR by collecting extra descriptions for each annotated moment. 
This dataset, named TV show Captions (\textbf{TVC}), is a large-scale multimodal video captioning dataset.
Fig.~\ref{fig:tvc_examples} shows two TVC examples.
Similar to TVR, the TVC task requires systems to gather information from both video and subtitle to generate relevant descriptions.
In the following, we present a brief analysis and initial baselines for TVC.

\begin{table}[!t]
\setlength{\tabcolsep}{0.15em}
\small
\centering
\small
\caption{Comparison of TVC with existing video captioning datasets. \textit{Desc. context} = Description context, it indicates which modality the descriptions are related to}
\scalebox{0.8}{
\begin{tabular}{p{3cm}P{1.5cm}P{2cm}P{1.2cm}P{2.5cm}P{1.1cm}P{1.1cm}P{2cm}}
\toprule
\multirow{2}{*}{\textbf{Dataset}} & \multirow{2}{*}{\textbf{Domain}} &\multirow{2}{*}{ \textbf{\#Moment}} & \multirow{2}{*}{\textbf{\#Desc.}} & \textbf{\#Desc. per} & \multicolumn{2}{c}{\textbf{Desc. context}} & \textbf{Desc. type} \\
 &  &  &  & \textbf{moment} & video & text & \textbf{anno.}\\
\midrule
TACoS-MLevel~\cite{rohrbach2014coherent} & Cooking & 25K & 75K & 3 & \checkmark & - & - \\
YouCook II~\cite{zhou2018towards} & Cooking & 15.4K & 15.4K & 11 & \checkmark & - & - \\
ANetCap~\cite{Krishna2017DenseCaptioningEI} & Activity & 100K & 100K & 1 & \checkmark & - & - \\
Charades~\cite{sigurdsson2016hollywood} & Indoor & 10K & 27.8K & 2-3 & \checkmark & - & - \\
VATEX~\cite{wang2019vatex} & Activity & 41K & 826K & 20 & \checkmark & - & - \\
LSMDC~\cite{rohrbach2017movie} & Movie & 128K & 128K & 1 & \checkmark & - & - \\
MST-VTT~\cite{xu2016msr} & Open & 10k & 200k & 20 & \checkmark & - & - \\
\midrule
TVC & TV show & 108K & 262K & 2-4 & \checkmark & \checkmark & \checkmark \\
\bottomrule
\end{tabular}
}
\label{tab:tvc_dset_comparison}
\end{table}

\begin{figure}[!t]
  \centering
  \includegraphics[width=0.9\linewidth]{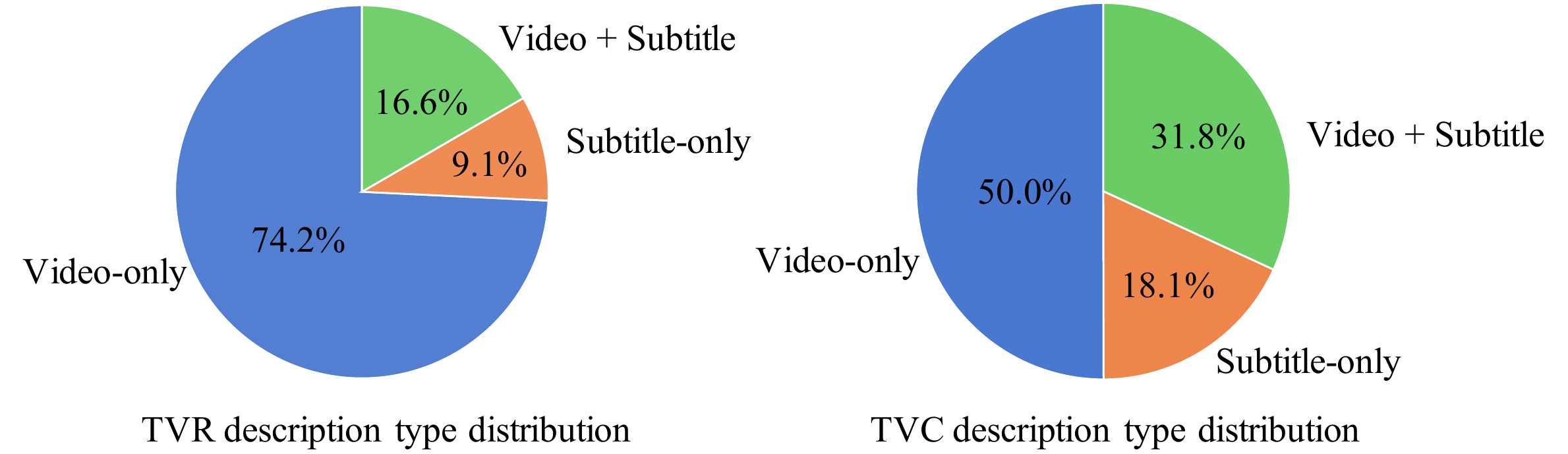}
  \caption{Description type distributions of TVR and TVC}
  \label{fig:tvc_tvr_qtype_dist}
\end{figure}

\subsection{Data Collection and Analysis}
To promote better coverage of the video (subtitle) content, we encourage annotators to write descriptions that are of different types from existing ones, e.g., we encourage annotators to write \textit{video-only} and \textit{video+sub} type descriptions if there already exists a \textit{sub-only} description. 
For each moment in the TVR training set, we collect one extra description, together with the original description forms the TVC training set with 2 descriptions for each moment.
For each moment in TVR val/test sets, we collect 4 extra descriptions as the TVC val/test sets.
The original val/test descriptions in TVR are not used to ensure data integrity. 
Details regarding data split are presented in Sec.~\ref{sec:data_release}.

Table~\ref{tab:tvc_dset_comparison} gives an overview of TVC and its comparison with recent video captioning datasets. 
In total, TVC contains 262K descriptions paired with 108K moments. 
TVC is unique as its captions may also describe dialogues/subtitles while the captions in the other datasets are only describing the visual content.
TVC also has a description type annotation, which can be used for model training and analysis.
Fig.~\ref{fig:tvc_tvr_qtype_dist} compares the description type distribution between TVR and TVC. 
As we encouraged annotators to write different types of descriptions, the description type distribution is more balanced in TVC compared to that of TVR. 
% Fig.~\ref{fig:tvc_sen_lengths} shows the description length distribution of TVC, the average length is 14.4 words (13.4 for TVR), median length is 13 words (12 for TVR). 
As TVC is built on top of TVR, it shares many properties of TVR, e.g., great linguistic diversity, rich inter-human interactions, more actions and people in a single description, etc. See Sec.~\ref{sec:dataset} for more details.

\begin{figure*}[!t]
  \centering
  \includegraphics[width=0.8\linewidth]{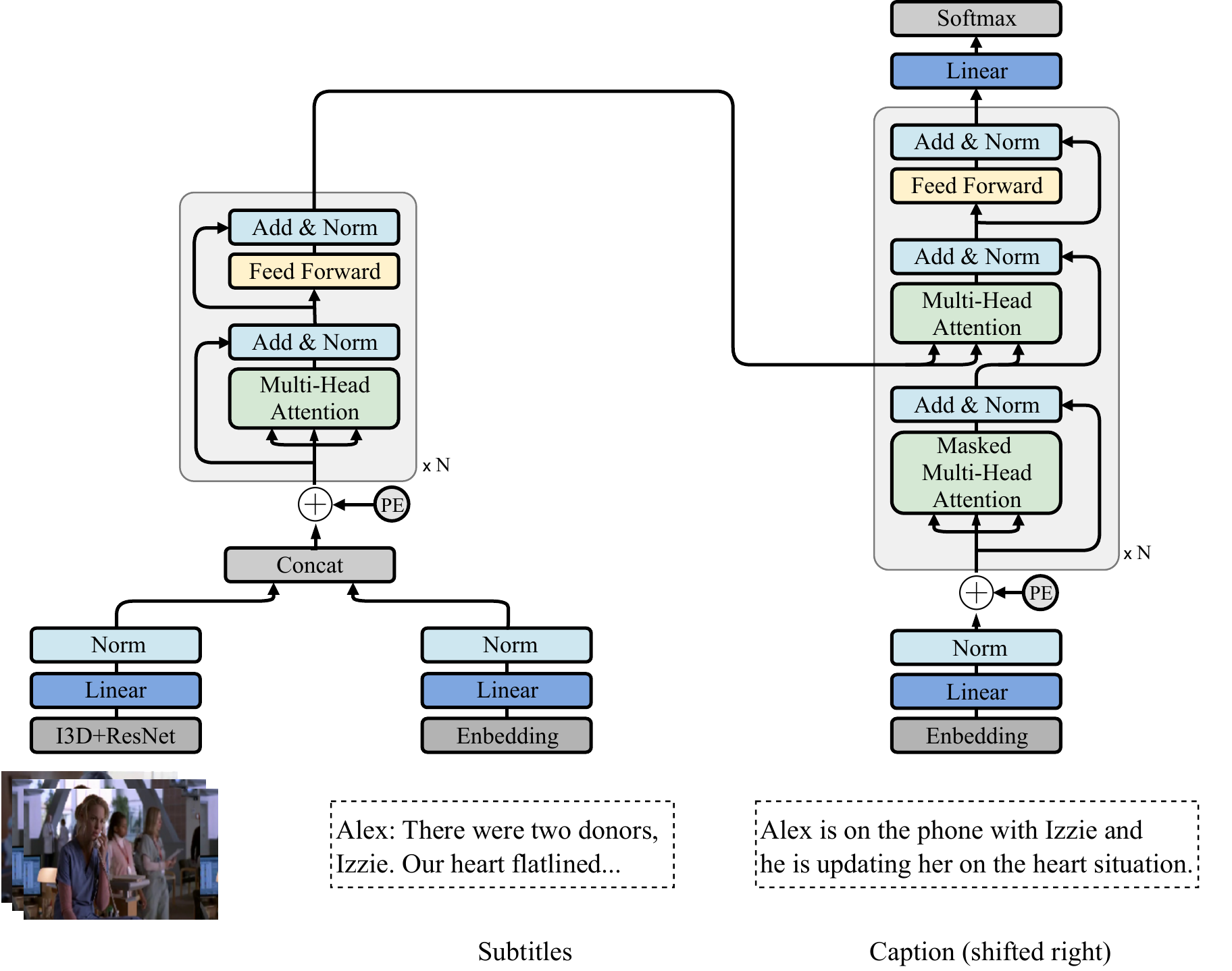}
  \caption{Overview of the MultiModal Transformer (MMT) model for the TVC task. \textit{PE} stands for Positional Encoding}
  \label{fig:tvc_model_overview}
\end{figure*}

\subsection{Multimodal Transformer}
To provide a strong initial baseline for the TVC multimodal video captioning task, we designed a MultiModal Transformer (\textbf{MMT}) captioning model which follows the classical encoder-decoder transformer architecture~\cite{vaswani2017attention}. 
It takes both video and subtitle as encoder inputs to generate the captions from the decoder.
Fig.~\ref{fig:tvc_model_overview} gives an overview of the designed model.

\kern2mm
\noindent\textbf{Input Representation}.
We use the concatenation of I3D~\cite{carreira2017quo} feature and ResNet-152~\cite{he2016deep} feature to represent videos. The features are pre-processed in the same way as our XML model for the TVR task, as in Sec.~\ref{subsec:xml_backbone}.
To represent subtitles, we use trainable 300D word embeddings. 
Next, we project raw video features and subtitle word features into a common embedding space using linear layers and layernorm~\cite{ba2016layer} layers. 
The projected video embedding $E^v \in \mathcal{R}^{l_{v}\times d}$ and subtitle embedding $E^s \in \mathcal{R}^{l_{s}\times d}$ are then concatenated at length dimension~\cite{lei2020mart} as the input to the encoder: $E^{ctx} = [E^v; E^s]$, where $E^{ctx} \in \mathcal{R}^{(l_v+l_s) \times d}$ stands for the context embedding, $d$ is hidden size.

\kern2mm
\noindent\textbf{Encoder and Decoder}. Both the encoder and decoder follows the standard design~\cite{vaswani2017attention} with 2 layers, i.e., $N \mbox{=} 2$. 
The decoder access encoder outputs at each layer with a multi-head attention~\cite{vaswani2017attention}.  
We refer readers to~\cite{vaswani2017attention} for a more detailed explanation of the model architecture.

\kern2mm
\noindent\textbf{Training and Inference}. 
We train the model using Maximum Likelihood Estimation (MLE), i.e., we maximize the likelihood of generating the ground truth words.
At inference, we use greedy decoding instead of beam search as it performs better in our experiments.

\begin{table}[!t]
\setlength{\tabcolsep}{0.3em}
\small
\centering
\small
\caption{Model comparison on TVC \textit{test-public} set, with different input context}
\scalebox{0.9}{
\begin{tabular}{p{3cm}
>{\raggedleft\arraybackslash}p{1.5cm}%
>{\raggedleft\arraybackslash}p{1.5cm}%
>{\raggedleft\arraybackslash}p{1.5cm}%
>{\raggedleft\arraybackslash}p{1.5cm}%
}
\toprule
Model & B@4 & METEOR & Rouge-L & CIDEr-D \\
\midrule
MMT (sub) & 6.33 & 13.92 & 7.73 & 33.76 \\
MMT (video) & 9.98 & 15.23 & 30.44 & 36.07 \\
MMT (video+sub) & \bf 10.87 & \bf 16.91 & \bf 32.81 & \bf 45.38 \\
\bottomrule
\end{tabular}
}
\label{tab:tvc_res}
\end{table}

\begin{table}[!t]
\setlength{\tabcolsep}{0.3em}
\centering
\small
\caption{Feature ablation on TVC \textit{val} set. All the models use both videos and subtitles}
\scalebox{0.9}{
\begin{tabular}{p{3cm}
>{\raggedleft\arraybackslash}p{1.5cm}%
>{\raggedleft\arraybackslash}p{1.5cm}%
>{\raggedleft\arraybackslash}p{1.5cm}%
>{\raggedleft\arraybackslash}p{1.5cm}%
}
\toprule
Model & B@4 & METEOR & Rouge-L & CIDEr-D \\
\midrule
MMT (ResNet) & 9.92 & 16.24 & 31.76 & 43.94 \\
MMT (I3D) & 10.25 & 16.48 & 31.98 & 43.70 \\
MMT (ResNet+I3D) & \bf 10.53 & \bf 16.61 & \bf 32.35 & \bf 44.39 \\
\bottomrule
\end{tabular}
}
\label{tab:tvc_feature_ablation}
\end{table}

\subsection{Experiments}
We use the same video split for TVC as in TVR, see Sec.~\ref{sec:data_release} for more details. We report numbers on standard metrics, inlcuding BLEU@4~\cite{papineni2002bleu},  METEOR~\cite{denkowski2014meteor}, Rouge-L~\cite{lin2004rouge}, CIDEr-D~\cite{vedantam2015cider}. 
We first compare MMT models with different input modalities. 
The results are shown in Table~\ref{tab:tvc_res}.
Across all metrics, the model with both videos and subtitles performs better than the models with only one of them, which shows both videos and subtitles are important for describing the moments.
Next, we compare models with different visual features. 
The results are shown in Table~\ref{tab:tvc_feature_ablation}. 
Models with both appearance features (ResNet-152~\cite{he2016deep}) and motion feature (I3D~\cite{carreira2017quo}) performs better than only using one of them.

\subsection{Qualitative Examples}
\begin{figure*}[!t]
  \centering
  \includegraphics[width=\linewidth]{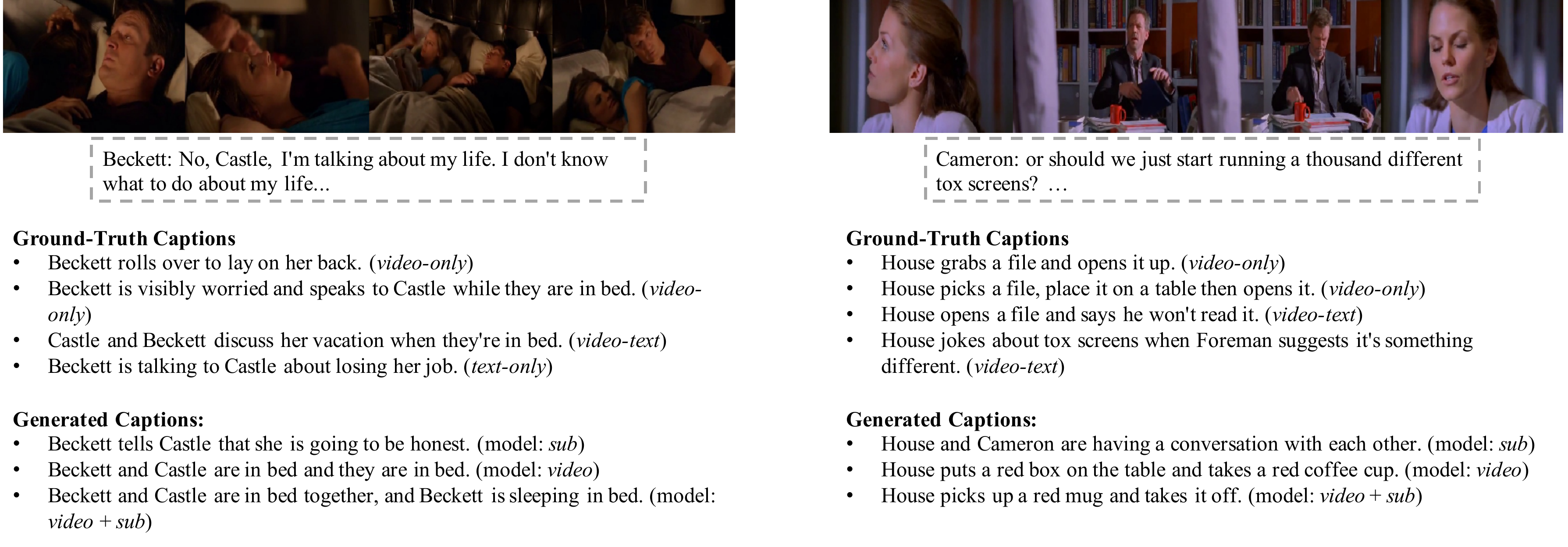} \\
  \includegraphics[width=\linewidth]{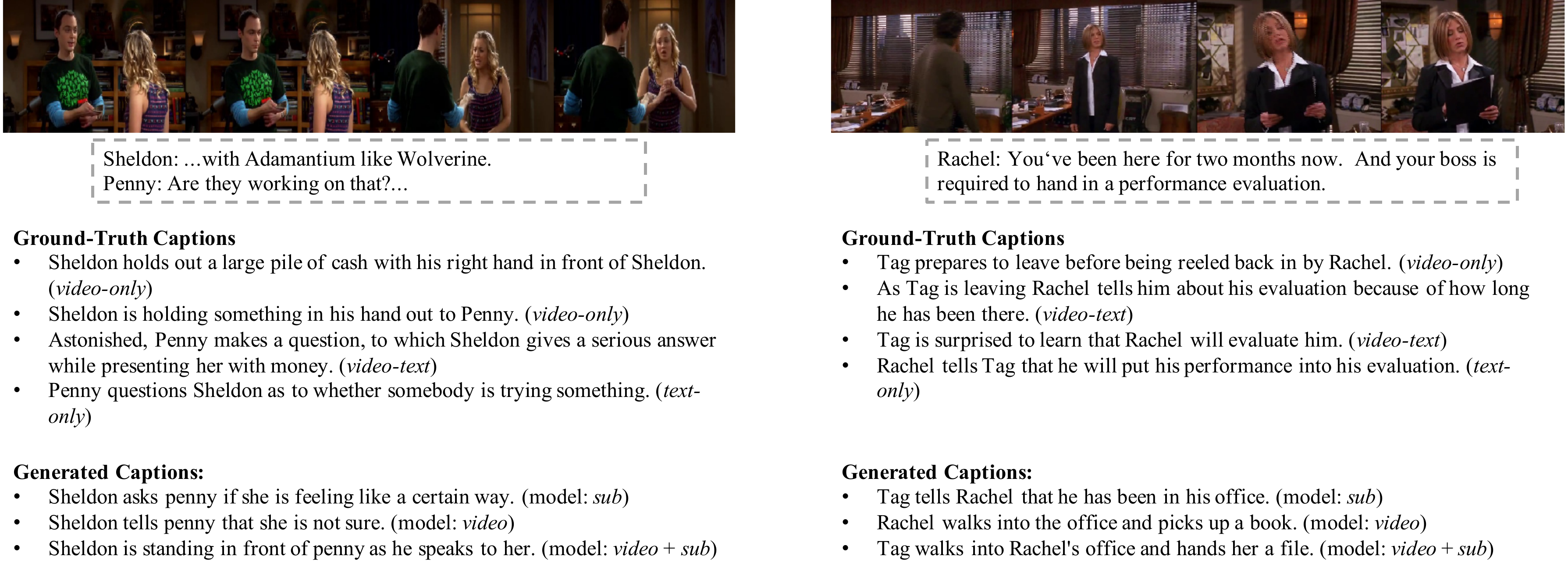} \\
  \includegraphics[width=\linewidth]{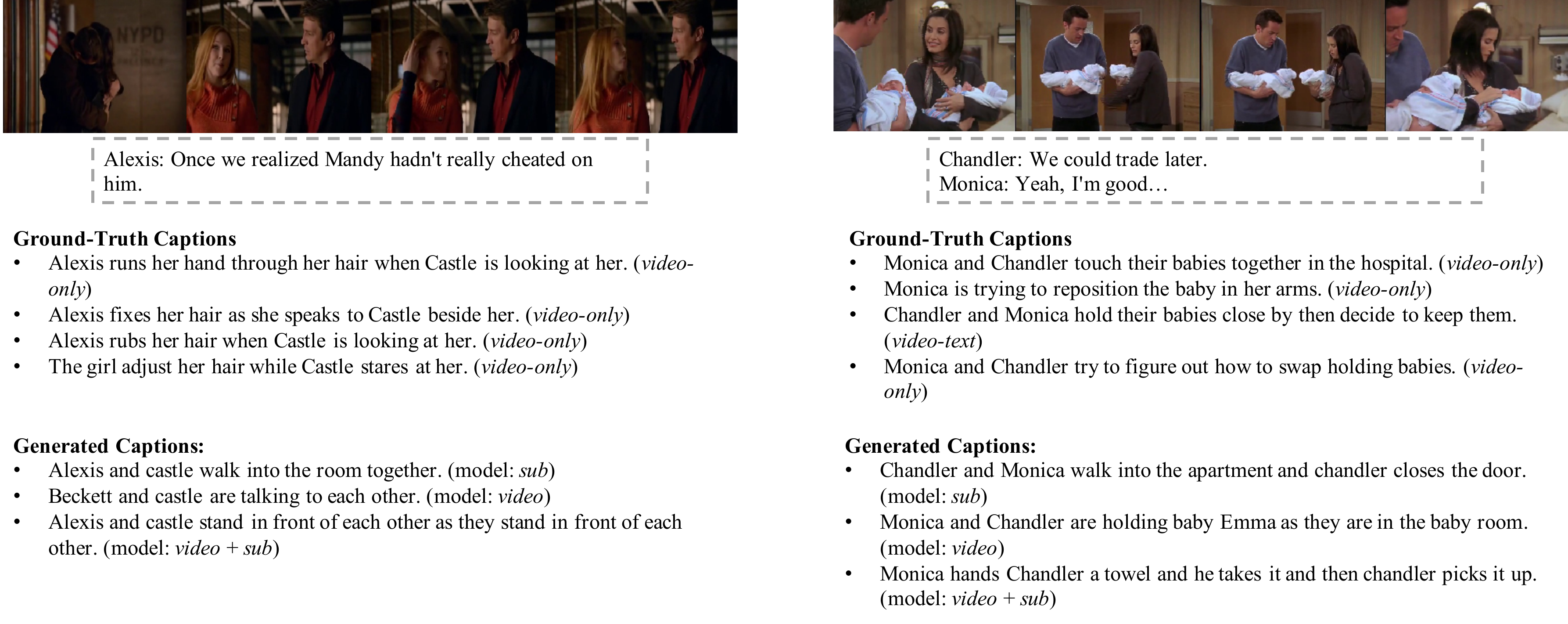}
  \caption{Qualitative comparison of MMT. Text inside \textit{dashed boxes} is the subtitles associated with the moments. Each ground-truth caption description is followed by a description type tag. We show comparison among models trained with only videos (\textit{video}), subtitles (\textit{sub}), or both (\textit{video} + \textit{sub})}
  \label{fig:tvc_model_prediction}
\end{figure*}

We show qualitative examples of MMT in Fig.~\ref{fig:tvc_model_prediction}, with generated captions by the three MMT models trained with different input context.

\section{Data Release and Public Leaderboards}\label{sec:data_release}
Both TVR and TVC are publicly available at their websites: \url{https://tvr.cs.unc.edu}, \url{https://tvr.cs.unc.edu/tvc.html}.
With the datasets, we host public leaderboards to better compare the systems.
In the following, we describe data split and usage in detail.

We split TVR into 80\% \textit{train}, 10\% \textit{val}, 5\% \textit{test-public} and 5\% \textit{test-private} such that videos and their associated queries appear in only one split.
This setup is the same as TVQA~\cite{Lei2018TVQALC}.
Details of the splits are presented in Table~\ref{tab:tvr_split}.
\textit{test-public} will be used for a public leaderboard, \textit{test-private} is reserved for future challenges. 
\textit{val} set should only be used for parameter tuning, it should not be used in the training process, including but not limited to pre-train the language features.

TVC follows the same data split as TVR, but with a different number of descriptions per moment, i.e., each of the training moments are paired with 2 descriptions while each of the moments in other splits are paired with 4 descriptions.
Details are presented in Table~\ref{tab:tvc_split}.
The rules on split usage are also the same as TVR.

\begin{table}[!t]
\setlength{\tabcolsep}{0.3em}
\centering
\small
\caption{TVR data split detail}
\scalebox{0.9}{
\begin{tabular}{p{2cm}
>{\raggedleft\arraybackslash}p{1.8cm}
>{\raggedleft\arraybackslash}p{1.8cm}
>{\raggedleft\arraybackslash}p{1.8cm}
}
\toprule
Split &  \#queries & \#moments  & \#videos\\
\midrule
\textit{train} & 87,175 & 87,175 & 17,435 \\
\textit{val} & 10,895 & 10,895 & 2,179 \\
\textit{test-public} & 5,445 & 5,445 & 1,089 \\
\textit{test-private} & 5,450 & 5,450 & 1,090 \\
\midrule
\textit{total} & 108,965 & 108,965 & 21,793 \\
\bottomrule
\end{tabular}
}
\label{tab:tvr_split}
\end{table}

\begin{table}[!t]
\setlength{\tabcolsep}{0.3em}
\centering
\small
\caption{TVC data split detail}
\scalebox{0.9}{
\begin{tabular}{p{2cm}
>{\raggedleft\arraybackslash}p{1.8cm}%
>{\raggedleft\arraybackslash}p{1.8cm}%
>{\raggedleft\arraybackslash}p{1.8cm}%
P{2.5cm}%
}
\toprule
Split & \#desc. & \#moments & \#videos & \#desc./moment \\
\midrule
\textit{train} & 174,350 & 86,603 & 17,435 & 2 \\
\textit{val} & 43,580 & 10,481 & 2,179 & 4 \\
\textit{test-public} & 21,780 & 5,420 & 1,089 & 4 \\
\textit{test-private} & 21,800 & 5,422 & 1,090 & 4 \\
\midrule
\textit{total} & 261,510 & 107,926 & 21,793 & - \\
\bottomrule
\end{tabular}
}
\label{tab:tvc_split}
\end{table}

\clearpage
% ---- Bibliography ----
%
% BibTeX users should specify bibliography style 'splncs04'.
% References will then be sorted and formatted in the correct style.
%
\bibliographystyle{splncs04}
\bibliography{egbib}
\end{document}